\documentclass[journal]{IEEEtran}
\ifCLASSINFOpdf
\else
\fi
\usepackage{graphicx}
\graphicspath{{./}{./assets/}{./photos/}}

\usepackage{booktabs,tabulary,multirow}
\usepackage[caption=false]{subfig}
 
\newcommand{\bd}[1]{\textbf{#1}}
\newcommand{\app}{\raise.17ex\hbox{$\scriptstyle\sim$}}

\def\x{\times}
\newcolumntype{x}[1]{>{\centering\arraybackslash}p{#1pt}}
\newcommand{\dt}[1]{\fontsize{8pt}{.1em}\selectfont \emph{#1}}
\newlength\savewidth\newcommand\shline{\noalign{\global\savewidth\arrayrulewidth
  \global\arrayrulewidth 1pt}\hline\noalign{\global\arrayrulewidth\savewidth}}
\newcommand{\tablestyle}[2]{\setlength{\tabcolsep}{#1}\renewcommand{\arraystretch}{#2}\centering\footnotesize}
\makeatletter\renewcommand\paragraph{\@startsection{paragraph}{4}{\z@}
  {.5em \@plus1ex \@minus.2ex}{-.5em}{\normalfont\normalsize\bfseries}}\makeatother

\usepackage{amsmath}
\usepackage{amssymb}
\usepackage{color,soul} % soul for highlighting text
\usepackage[table]{xcolor}
\definecolor{lightgray}{gray}{0.9}
\definecolor{citecolor}{RGB}{34,139,34}
\usepackage[pagebackref=true,breaklinks=true,letterpaper=true,colorlinks,
  citecolor=citecolor,bookmarks=false]{hyperref}

\def\bx{{\bf x}}
\def\by{{\bf y}}

\def\vs{{\textit{vs.~}}}

\usepackage{bm}

\providecommand{\newoperator}[3]{%
\newcommand*{#1}{\mathop{#2}#3}}
\newoperator{\argmin}{\mathrm{argmin}}{\limits}
\newoperator{\argmax}{\mathrm{argmax}}{\limits}

\usepackage{pifont}% http://ctan.org/pkg/pifont
\newcommand{\cmark}{\ding{51}}%
\newcommand{\xmark}{\ding{55}}%

\usepackage{overpic}
\begin{document}
\title{Learning a Layout Transfer Network for Context Aware Object Detection}

\author{\IEEEauthorblockN{Tao Wang,
Xuming He,
Yuanzheng Cai,
and Guobao Xiao}

\thanks{T. Wang, Y. Cai and G. Xiao are with the Fujian Provincial Key Laboratory
of Information Processing and Intelligent Control, Minjiang University, Fuzhou,
China.}% <-this % stops a space
\thanks{X. He is with the School of Information Science and Technology,
ShanghaiTech University, Shanghai, China.}
\thanks{T. Wang and Y. Cai are also with the College of Mathematics and Computer Science,
Fuzhou University, Fuzhou, China and NetDragon Inc., Fuzhou, China.}
\thanks{Digital Object Identifier 10.1109/TITS.2019.2939213}
% <-this % stops a space
}


\markboth{IEEE TRANSACTIONS ON INTELLIGENT TRANSPORTATION SYSTEMS}%
{Shell \MakeLowercase{\textit{et al.}}: Bare Demo of IEEEtran.cls for IEEE Journals}



\maketitle

\begin{abstract}
We present a context aware object detection method based on a
retrieve-and-transform scene layout model.
Given an input image, our approach first retrieves a coarse scene layout
from a codebook of typical layout templates.
In order to handle large layout variations, we use a variant of the
spatial transformer network~\cite{jaderberg2015spatial} to transform and refine the retrieved layout, resulting in
a set of interpretable and semantically meaningful feature maps of object locations and scales.
The above steps are implemented as a Layout
Transfer Network which we integrate into Faster RCNN~\cite{ren2015faster} to allow for joint
reasoning of object detection and scene layout estimation.
Extensive experiments on three public datasets verified that our approach
provides consistent performance improvements to the state-of-the-art
object detection baselines on a variety of challenging
tasks in the traffic surveillance and the autonomous driving domains.
\end{abstract}


% \fi
\IEEEpeerreviewmaketitle

\section{Introduction}
\label{sec:intro}
\begin{figure}
      \centering
		\includegraphics[width=0.45\textwidth]{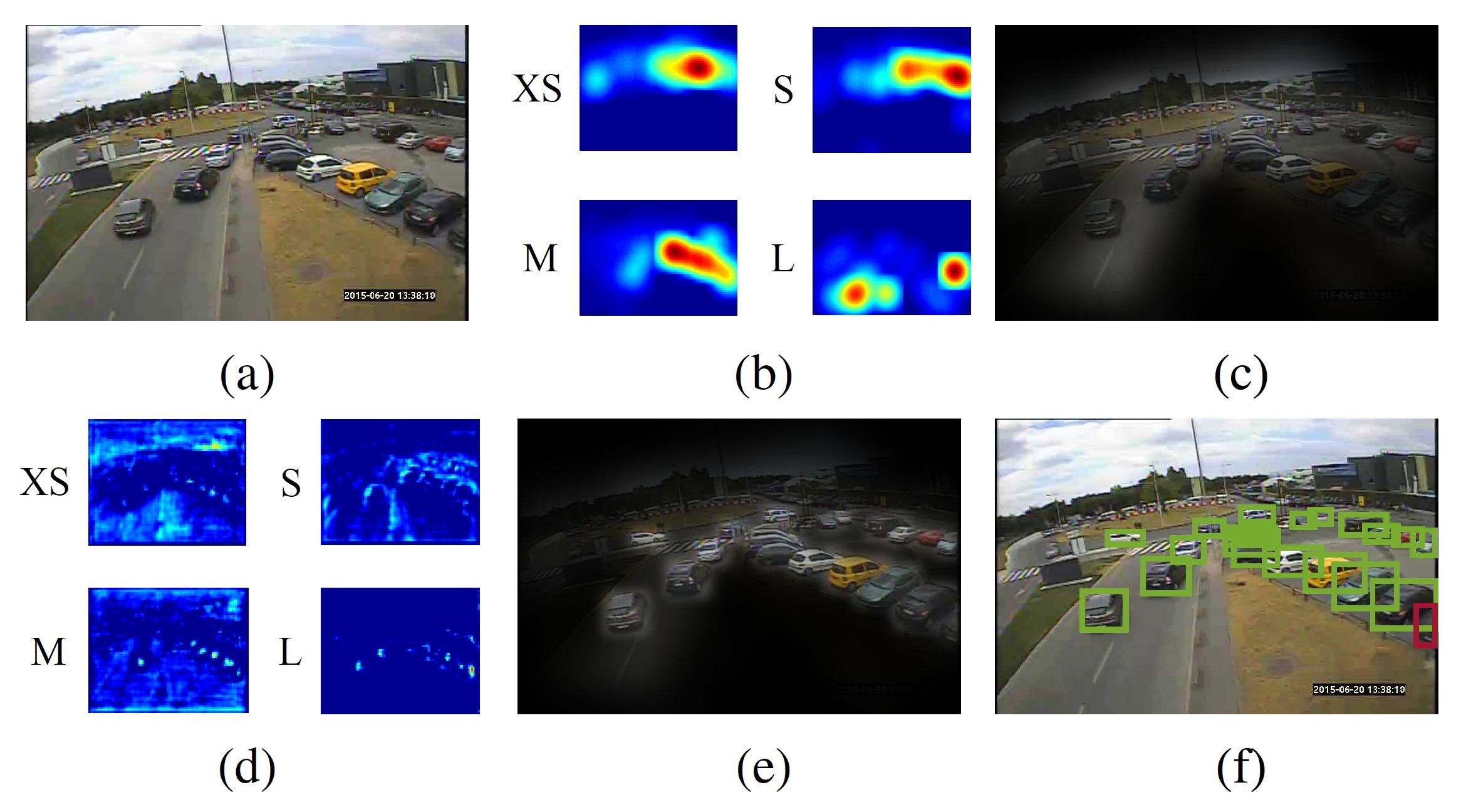}
      \caption{Transferring scene layouts for object detection. \textbf{(a)} Input image.
		    \textbf{(b)} Coarse scene layout for the \textit{car} category at $4$ different scales (XS, S, M, L) retrieved from similar images.
            \textbf{(c)} Object attention map derived from the coarse scene layout.
            \textbf{(d)} Scene layout after transformation and refinement.
            \textbf{(e)} Object attention map derived from the transformed and refined scene layout.
            \textbf{(f)} Detector output.}
      \label{fig:motivation}
      \vspace{-2mm}
\end{figure}

\IEEEPARstart{H}{uman} perception almost always reflects an integration of
information from bottom-up and top-down processes (e.g.,~\cite{mcclelland1981interactive,palmer1999vision,lee2003hierarchical}).
In particular, context effects are an integral part of visual perception in humans.
Consider the object detection problem as shown in Figure~\ref{fig:motivation}.
As humans, we take for granted the ability to understand the layout of a scene at the first glance,
and then know where to look for specific objects.
For example, cars are most likely to appear on paved areas and those closer to the camera are typically larger than distant ones.
It is therefore appealing to design models for object detection that reason about the
spatial context in a similar fashion.
On the contrary, recent state-of-the-art object detection algorithms
produce scores for densely sampled object locations and scales,
or a few hundreds to thousands of ``blobby'' object proposals.
While it is true that these methods encode the spatial context to some extent due to the large
receptive fields of the underlying convolutional neural nets (CNNs), methods like these usually lack a holistic
understanding of the scene layouts.
More importantly, unlike human perception, their performances deteriorate quickly when
visual cues of objects become ambiguous or weak.
Motivated by how humans understand the scene layouts for object detection, we propose a
simple, interpretable and flexible framework for learning to transfer scene layouts for context-aware object detection.

\begin{figure*}[ht!]
	\begin{center}
		\includegraphics[width=0.9\textwidth]{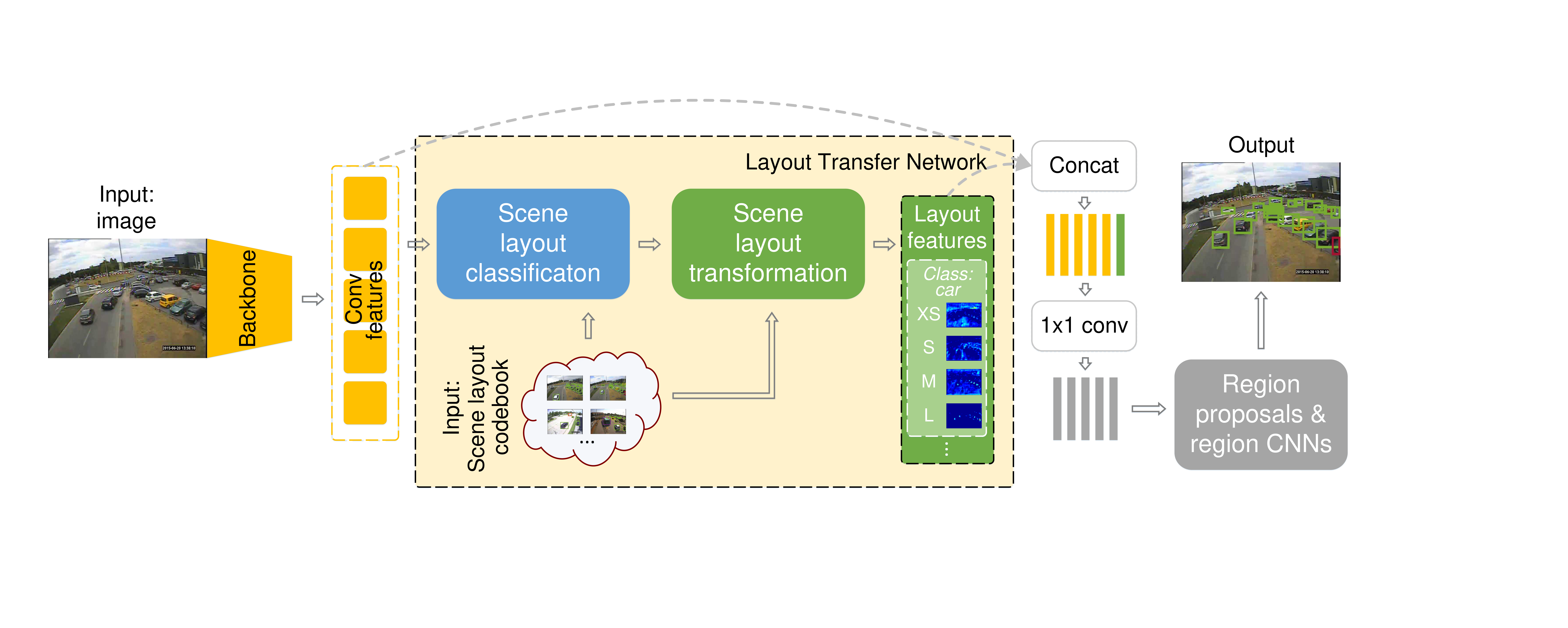}
	\end{center}
	\caption{The overall architecture of our method. Our layout transfer network adopts a retrieve-and-transform strategy for
	scene layout estimation. Given an input image, we first retrieve the most similar scene layout from a pre-built codebook
	with a classification sub-network (blue). The retrieved layout are then refined with a transformation sub-network, producing
	the final scene layout features (green). These features are then concatenated with backbone convolutional features (orange) to obtain
	the feature map for downstream object detection (gray).
	}
	\label{fig:net-struct}
\end{figure*}

The general idea of context modeling for object detection has long been proven effective in the computer vision community,
with seminal works from Torralba et al.~\cite{torralba2003contextual,murphy2003using,torralba2003context}, and later Hoiem, Efros and Hebert~\cite{hoiem2008putting},
plus a few more~\cite{wolf2006critical,rabinovich2007objects,blaschko2009object} as prominent examples.
More recently, the modeling of the spatial context has been extended to 3D
scenarios~\cite{sudderth2006depth, Bao:CVPR10, sun2010object,choi2013understanding, lin2013holistic,gupta2014learning, liu2015towards} as
high quality co-registered depth and color images have become more easily accessible.
Most existing approaches assume a parametric model of the scene layout, such as the piecewise planar assumption~\cite{faugeras1988motion},
blocks world assumption~\cite{gupta2010blocks},
or the Manhattan world assumption~\cite{hedau2009recovering,lee2009geometric,lee2010estimating,chao2013layout}, just to name a few.
Despite the great progress, modeling the scene layout in a parametric fashion becomes challenging when the layout variation is large.
In addition to an increased model complexity, performance could deterioriate quickly under atypical scene configurations.
To address the above issues, we explore a semi-parametric approach to context modeling for object detection.
In particular, we improve object detection through a coarse-to-fine scene layout model that predicts potential object locations and scales
for every object category, as illustrated in Figure~\ref{fig:motivation}.
Such a high dimensional mapping is difficult to learn in a purely parametric way due to the large output space,
so we adopt a retrieve-and-transform strategy for scene layout prediction.
Specifically, the input image (Figure~\ref{fig:motivation}~(a)) is first matched against a codebook of typical scene layout templates,
and the most similar scene layout (Figure~\ref{fig:motivation}~(b) and (c)) is retrieved.
This scene layout is further transformed and refined (Figure~\ref{fig:motivation} (d) and (e)) to adapt to the specific appearance of the input image.
In addition, we implement the above procedure as an integral component of Faster RCNN~\cite{ren2015faster}, so that the object detection and the scene layout estimation tasks can be learned
in an alternating fashion.

We note that the proposed method resembles human perception in the sense that we are able to seamlessly integrate
information from bottom-up and top-down visual processings.
The bottom-up processing involves the process that starts with deep
image features; the top-down processing, which starts with retrieving scene layouts from our external scene layout memory, injects
prior knowledge and expectations about the scene.
By virtue of a simple and differentiable transformation, the proposed method is able to adapt the
most relevant scene layout priors to a specific input image.

The benefit of our proposed method is threefold.
Firstly, our scene layout representation is interpretable, which makes it more readily applicable to other
scene understanding tasks. In fact, we show that
we are able to build an external memory of typical scene layouts from a
large database and then accurately retrieve the most relevant scene templates at test time.
Secondly, a common problem in these retrieval-based methods is that it would be challenging to 
deal with intra-class variations. In our work, we use a variant of the spatial transformer network~\cite{jaderberg2015spatial}
to adapt scene layout templates to specific images, making our method capable of handling diverse scene layouts.
Lastly, the proposed module can be integrated into a deep network for object detection,
resulting in a single CNN for joint object detection and scene layout estimation.
Extensive experiments on three public datasets verified that images in the traffic surveillance and automated driving domains
are well-suited for our approach because their scene layouts provide strong priors for localizing objects.

This paper extends our previous work~\cite{wang2017efficient} in several ways. In particular, (i) scene layout estimation
is now based on a drastically different method based on retrieve-and-transform, which is both more efficient and effective;
(ii) scene layout estimation is now an integral part of the object detection CNN. This improves detection
performance, and also removes unnecessary feature computational costs; (iii) we provide more detailed experimental evaluation
and ablation studies on our proposed method in order to quantitatively motivate our model design. Furthermore, we compare
our method against the baselines on two additional public datasets for object detection.

The rest of the paper is organized as follows. In Section~\ref{sec:related}, we briefly discuss related work in object detection, context modeling,
and scene layout estimation.
The details of our model, including its structure, inference and learning, are introduced in Section~\ref{sec:approach}.
This is followed by experimental evaluation in Section~\ref{sec:experiments} and closing remarks in Section~\ref{sec:conclusion}.

\section{Related Work}
\label{sec:related}

\noindent \textbf{Object detection.} Recent years witnessed a huge success of Convolutional Neural Network (CNN) based
object detection algorithms over conventional methods
based on hand-crafted features and a shallow object grammar-based architecture.
Some of the most prominent examples include sliding-window based OverFeat~\cite{sermanet2014overfeat} and object proposal based R-CNN~\cite{girshick2014rich} and its faster variants~\cite{he2014spatial,girshick2015fast,ren2015faster,dai2016r}.
These methods are directly inspired by the success of CNN for image classification. The latter, proposal-based methods seek to exploit the strong representation power
of deep networks to classify and make refinements to a relatively small set (typically hundreds to a few thousands) of potential object regions.
Another line of work attempts to make direct predictions using a deep network without the object proposal step. Examples include
YOLO~\cite{redmon2016you}, SSD~\cite{liu2016ssd}, DSOD~\cite{shen2017dsod} and
we note that these methods are generally more computationally efficient.
In this work, we choose Faster RCNN~\cite{ren2015faster} as our baseline object detector and explore how to improve their results
via incorporating scene-level context cues.

\noindent \textbf{Context modeling.} Context aware object detection has been well studied, and many context-aware object detection methods have been proposed
(e.g.,~\cite{torralba2003contextual,torralba2004contextual,wolf2006critical,rabinovich2007objects,hoiem2008putting,kluckner2009semantic,blaschko2009object,maire2011object,pan2013coherent}).
See~\cite{wolf2006critical} for a review and~\cite{divvala2009empirical} for an empirical study of earlier work in the literature.
More recently, Yang et al.~\cite{yang2010layered} have shown that reasoning about a 2.1D 
layered object representation in a scene can positively impact object detection.
Yao et al.~\cite{yao2012describing} propose a holistic scene understanding model which jointly 
solves object detection, segmentation and scene classification. 
Mottaghi et al.~\cite{mottaghi2014role} exploit both the local and global contexts by reasoning about the presence of contextual classes, and
propose a context-aware improvement to the DPM~\cite{felzenszwalb2010object}.
Zhang et al.~\cite{zhang2014data} propose a nonparametric column-based tiered model for scene layout estimation of road scenes.
Zhu et al.~\cite{zhu2015segdeepm} use CNNs to obtain contextual scores for object hypotheses, in addition to scores obtained with object appearance.
Cai et al.~\cite{cai2016unified} propose to ease the object scale variation issue by performing object detection with multiple CNN layers, each
focusing on objects within certain scale ranges.
Batzer et al.~\cite{batzer2016generic} propose a context-aware voting scheme for small and distant object detection.
In addition, Sun and Jacobs~\cite{sun2017seeing} propose to learn a context model that predicts where objects may be missing.
Other works have extended context modeling to 3D scenarios.
For example, Bao, Sun and Savarse propose a parameterized 3D surface layout model and combine it with object detectors~\cite{Bao:CVPR10, sun2010object}.
Geiger, Wojek and Urtasun~\cite{geiger2011joint} propose a generative model for joint inference of scene topology, geometry and 3D object locations.
Wojek et al.~\cite{wojek2013monocular} also propose a 3D scene model with explicit occlusion reasoning for object detection and tracking.
Choi et al.~\cite{choi2013understanding} learn latent 3D geometric phrases to jointly solve object detection and scene layout estimation.
Similarly, Lin et al.~\cite{lin2013holistic} use a CRF model to integrate various contextual relations for holistic scene understanding.
Other later works include~\cite{gupta2014learning},~\cite{wang2015holistic} and~\cite{liu2015towards}.
Our work differs from the methods above in the sense that we propose a semi-parametric, retrieve-and-transform based approach to model the
spatial context for object detection, which allows us to efficiently search within a high dimensional output space of the scene context.
Our method is simple, interpretable, and it can be integrated into a deep network for object detection.

\noindent \textbf{Scene layout estimation.} Our work is also related to scene layout estimation methods that attempt to predict
either parametric or nonparametric scene layout representations.
For indoor scenes, recent works (e.g.,~\cite{mallya2015learning,dasgupta2016delay,liu2015rent3d,armeni20163d,song2017semantic}) made great progress by
leveraging strong scene priors such as floor plans, geometric priors, or the more classical Manhattan world assumption, in conjunction with deep models trained on
large-scale datasets.
For outdoor scenes, Seff and Xiao~\cite{seff2016learning} propose to use deep models to predict a set of driving-related
road layout attributes.
In addition, Mattyus et al.~\cite{mattyus2016hd} propose a CRF-based scene layout model that combines perspective and top-view images.
Zhai et al.~\cite{zhai2017predicting} learn a transform to transfer the semantics from ground to the aerial image domain.
Li et al.~\cite{li2017foveanet} propose a perspective-aware scene parsing method that estimates the perspective
geometry of a scene image through a CNN to allow for finer parsing of small distant objects, and fuse prediction results at multiple scales with a perspective-aware CRF.
Schulter et al.~\cite{schulter2018learning} propose a CNN that can hallucinate depth and semantics occluded by foreground objects and
estimate a scene layout in the top view from a single perspective view.
Wang et al.~\cite{wang2019parametric} propose a rich parametric top-view representation of complex road scenes that uses CNNs for predicting scene model parameters
and a CRF for consistency reasoning.
Contrary to the above methods, we propose a simple and interpretable scene layout representation that can be
directly used to improve object detection performance, and does not require additional data or annotation during training.
Furthermore, we implement our retrieve-and-transform scene layout estimation model as part of an
object detector that allows jointly learning for object detection and scene layout estimation.

\begin{table}[t]
	\centering
    \footnotesize
			\begin{tabular}{l l}
             \multicolumn{2}{l}{} \\[-0.9em] % Spacer
             \multicolumn{2}{l}{Object detection} \\[.1em]
\hline
             \multicolumn{2}{l}{} \\[-0.9em] % Spacer
			 input image & $I$ \\
			 object hypothesis & $\bx=(\bx_c, a_s, a_r, o)$ \\
			 object detection score & $S(\bx|I)$ \\
\hline
             \multicolumn{2}{l}{} \\[-0.9em] % Spacer
             \multicolumn{2}{l}{Scene layout} \\[.1em]
\hline
             \multicolumn{2}{l}{} \\[-0.9em] % Spacer
			 training dataset & $\mathcal{D} = \{(I_m, \mathcal{Y}_m,z_m)\}_{m=1}^M$ \\             
             ground-truth annotation & $\by = (\by_c, a_s, a_r, o), \by \in \mathcal{Y}_m$ \\
             scene layout type & $z_m \in \mathcal{Z}, \mathcal{Z}=\{1,\cdots,N\}$ \\             
             image neighborhood & $\mathcal{N}_I, \mathcal{N}_I \in \mathcal{D}$ \\             
             scene layout codebook & $\mathcal{C}=\{(z_m, \mathcal{Y}_m)\}_{m=1}^M$ \\
             coarse scene layout score & $S_c(\bx|\mathcal{C})$ \\
             refined scene layout score & $S_l(\bx|\mathcal{N}_I) = \mathcal{T}\big(S_c(\bx|\mathcal{C})\big)$ \\
         \end{tabular}
         \vspace{2mm}
         \caption
         {
            Summary of the main notations. See Section~\ref{sec:approach} for details.
         }
\label{tab:notations}
\vspace{-2mm}

\end{table}

\section{Our Approach}
\label{sec:approach}

Let us begin with the definition of the object detection problem and notations.
Given an input image, object detection algorithms output a score for each valid object hypothesis.
More formally, suppose we have an input image $I \in \mathcal{I}$ where $\mathcal{I}$ is the input image space.
Let the object hypothesis be $\bx\in\mathcal{X}$, where $\mathcal{X}$ is the object hypothesis space.
To simplify the notation, we assume each hypothesis is $\bx=(\bx_c,a_s,a_r,o)$ where $\bx_c=(a_x,a_y)$ is the image coordinate
location of the object center, $a_s$ the scale, $a_r$ the aspect ratio, and $o \in \mathcal{O}$ the object class.
Note that each $\bx$ now implies a bounding box as well.
Object detection algorithms define a scoring function $S(\bx|I)$ for each valid object hypothesis $\bx$.
For example, this score is implemented as a two-class softmax score in Faster RCNN, i.e., $S(\bx|I)=p(\bx|I)$.

The core idea of this work is that we are able to transfer scene layouts from training images that are similar to the input image
to predict potential object locations and scales.
To this end, we propose an additional scene \textit{layout} score $S_l(\bx|\mathcal{N}_I)$ for any given object hypothesis $\bx$
by investigating a local neighborhood $\mathcal{N}_I$ of the input image $I$. %defined on an appearance feature manifold.
Directly learning such a high-dimensional mapping is difficult, so we adopt a two-step retrieve-and-transform strategy
here:
\begin{itemize}
\item We first retrieve a \textit{coarse} scene layout score $S_c(\bx|\mathcal{C})$
by matching the input image $I$ to a codebook $\mathcal{C}$ of scene layout \textit{templates}.

\item Afterwards, we apply a transformation $\mathcal{T}$ to $S_c(\cdot)$
to obtain the refined scene layout score, i.e., $S_l(\bx|\mathcal{N}_I) = \mathcal{T}\big(S_c(\bx|\mathcal{C})\big)$.
\end{itemize}

\begin{figure*}[ht]
	  \centering
	  \includegraphics[width=1.0\textwidth]{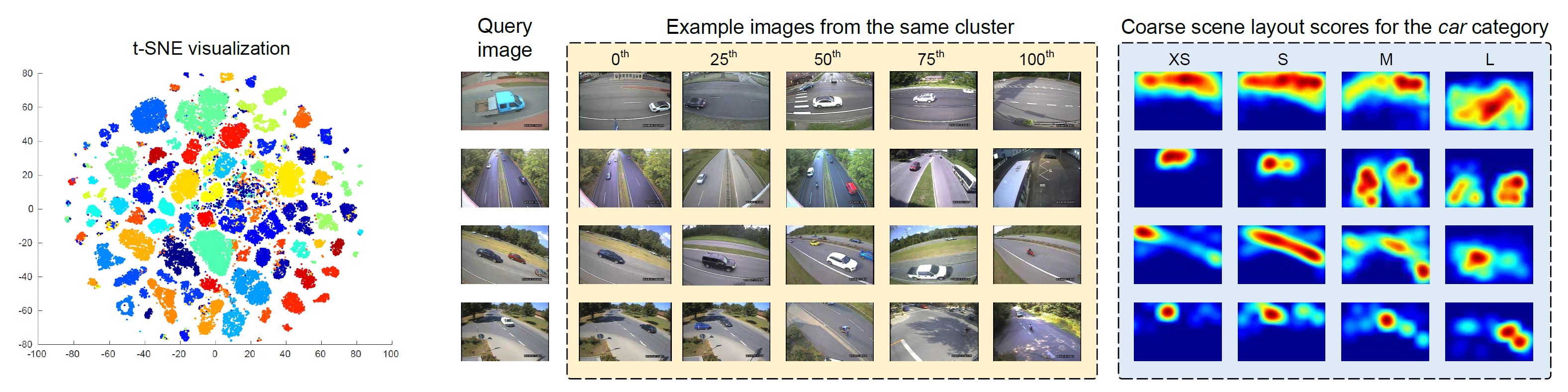}
		\caption
         {
         	\textbf{Left}: The t-SNE visualization~\cite{maaten2008visualizing}
         	of the $\mathit{pool5}$ features from the ResNet-50~\cite{he2016deep} network, color-coded
         	according to their cluster membership.
         	\textbf{Right}: Given a query image (in the leftmost column), we present example images
         	from the matching scene layout cluster on the MIO-TCD dataset (the yellow panel).
         	In particular, we show images near the 0th, 25th, 50th, 75th and 100th percentiles
         	according to their distances from the cluster center.
			In addition, we show coarse scene layout scores $S_c(\cdot)$ for the \textit{car} category obtained
			on the MIO-TCD dataset (the blue panel).
         	By matching a query image to our scene layout codebook, spatial distribution
         	of object locations at $4$ different scales (XS, S, M, L) are obtained.
         	Note that the scene layout scores are still coarse,
         	but they roughly suggest a scene layout in terms of potential object locations.
         }
\label{fig:codebook}
\end{figure*}

More specifically, denote the training dataset as $\mathcal{D} = \{(I_m, \mathcal{Y}_m,z_m)\}_{m=1}^M$ with $M$ images, where $I_m$
is a training image and $\mathcal{Y}_m$ is the set of all ground-truth annotations of the $m$-th image.
Similar to $\bx$, let $\by = (\by_c, a_s, a_r, o), \by \in \mathcal{Y}_m$ be a ground-truth object annotation.
Instead of codewords, our scene layout codebook $\mathcal{C}$ stores the scene layout \textit{type} label $z_m$ and
the ground-truth annotations of each training
image, i.e., $\mathcal{C}=\{(z_m, \mathcal{Y}_m)\}_{m=1}^M$, $z_m \in \mathcal{Z}$, where $\mathcal{Z}=\{1,\cdots,N\}$ and $N$ is the number of scene layout templates.
In this way, each training image $I_m$ is additionally associated with its scene layout type label $z_m$.
The main notations are summarized in Table~\ref{tab:notations}.

There are three key issues here: (1) how to define the neighborhood $\mathcal{N}_I$, (2) how to obtain an
appropriate representation for $S_l(\bx|\mathcal{N}_I)$ that can be helpful
for object detection, and (3) how to integrate $S_l(\bx|\mathcal{N}_I)$ into an existing object detection framework.
To address the first issue, we build a codebook of typical scene layouts by clustering the image-level appearance features.
By matching an input image against the codebook,
we can learn to transfer the neighborhood information encoded in the codebook entries to the target image.
See Figure~\ref{fig:codebook} for examples.
For the second issue, we choose the object location heatmaps, since our ultimate goal is to predict object locations.
In this work, we follow a retrieve-and-transform strategy to obtain a feature map that not only encodes the spatial distribution of objects in the
image neighborhood, but also further adapted to specific images based on the input image appearance.
The overall process is illustrated in Figure~\ref{fig:net-struct}, 
with examples of the output scene layout scores shown in Figure~\ref{fig:transformer-output}.
For the last issue, we propose two strategies for information fusion either at the feature level or at the final decision scores, respectively.

It should be noted that our method provides
interpretable intermediate features of the coarse layout $S_c(\cdot)$ and the refined layout $S_l(\cdot)$ that resemble the psychological
process of humans looking up the memory of typical
scene layouts, and then adapt them to specific scene appearances.
This also allows us to inject additional supervision signals to help convergence during training.
In addition, such a high-dimensional mapping is difficult to learn in a purely parametric way due to the large output space.
We quantitatively show in the experiments that our strategy
is much more effective than a naive baseline which attempts to directly predict the spatial locations of objects.

\subsection{Building a scene layout codebook}
\label{sec:codebook}

As the first foundation stone, we obtain scene layout templates in the training dataset by clustering the image-level appearance features.
We build a codebook $\mathcal{C}$ that encodes typical scene layouts,
and train a classifier for
the scene layout classification. At test time, we then classify the input image into one of these scene layout clusters, and obtain a rough estimate possible of object locations.
To this end, we introduce a feature manifold that is descriptive of the scene layouts.
Following~\cite{wang2017efficient}, we use the $2048$-D features extracted from the $\mathit{pool5}$ layer of a ResNet-50~\cite{he2016deep}
network applied on the input image $I$, as it has been widely used in image classification to describe the appearance of an image.
The network is pretrained on the ImageNet dataset~\cite{deng2009imagenet,he2016deep}.
One advantage of these features is that they are pretrained on a large database and are potentially a more robust
representation of the image appearance when compared to other alternatives.

More specifically, let $M$ be the number of images in our training set and $D$ be the feature dimension for
the image-level appearance features (i.e., $D=2048$ for our ResNet $\mathit{pool5}$ features).
We can perform K-means clustering with $N$ clusters for the
training feature matrix $\mathbf{F} \in \mathbb{R}^{D \times M}$.
Denote $z_m$ as the cluster membership for the $m$-th image in the training set,
and $\delta(z_m=i)$ an indicator function for whether or not the $m$-th image belongs to the $i$-th cluster.
Our scene layout codebook stores the cluster membership labels and the ground-truth object annotations
in a nonparametric fashion, i.e., $\mathcal{C} = \{(z_m, \mathcal{Y}_m)\}_{m=1}^M$.
The scene layout score for the $i$-th cluster $S_c^i(\bx|\mathcal{C})$ is obtained by accumulating all
ground-truth object annotations in that cluster:

\begin{align}
      S_c^i(\bx|\mathcal{C}) = \sum_{m=1}^{M}\frac{\delta(z_m=i)}{Z_c\sigma^2} \sum_{\substack{\bx \simeq \by\\
      \by \in \mathcal{Y}_m}} \exp \Big( - \frac{||\by_c - \bx_c||^2}{2 \sigma^2} \Big)
	  \label{eqn:vote}
\end{align}

\noindent where $\by = (\by_c, a_s, a_r, o), \by \in \mathcal{Y}_m$ is a ground-truth object annotation of the $m$-th image,
and $\mathcal{Y}_m$ is the set of all
ground-truth object annotations in the $m$-th image.
We build a mixture model with $K$ components by sorting $\by$'s by their scale $a_s$, aspect
ratio $a_r$, object class $o$ and split them into $K$ groups $Y_1,...,Y_K$.
Each of these groups includes annotations within certain ranges of $a_s$ and $a_r$, and of a specific object class $o$.
Similar to $\by$, an object hypothesis $\bx$ can be sorted into one of these groups.
For notation simplicity,
we use $\bx \simeq \by$ to denote that $\bx$ and $\by$ are in the same group.
This is straightforward because we would only like to accumulate support for an object hypothesis from those ground-truth 
annotations that are in the same group.
In addition, $\by_c$ and $\bx_c$ are the object center coordinates, and $Z_c$ is a normalizing factor.
Equation~\ref{eqn:vote} additively
accumulates support for an object hypothesis $\bx$ by looking at nearby spatial locations 
in the ground-truth object annotations while applying a Gaussian smoothing kernel.

The left part of Figure~\ref{fig:codebook} presents the t-SNE visualization~\cite{maaten2008visualizing} of the
training feature matrix $\mathbf{F}$, with each feature color-coded according to its
cluster membership. The right part of Figure~\ref{fig:codebook} shows a query image and example images (in the
yellow panel) from the matching scene layout cluster on the MIO-TCD dataset. In particular, we show images near
the 0th, 25th, 50th, 75th and 100th percentiles according to their distances from the cluster center. In this way,
we are able to show the variations within individual clusters. Specifically, a scene layout cluster contains images
taken with different cameras from similar views, not merely different images taken from the same camera.
In addition, we show the
spatial distribution of objects in the \textit{car} category at $4$ different scales
(in the blue panel).
We note that the heatmaps in Figure~\ref{fig:codebook} can be viewed as a sampling distribution of object locations
obtained from a specific image cluster, and they do not generally provide exact object locations for a given image.
The intra-cluster layout variations
are still large, and that we further refine the scene layout scores to 
adapt to the appearance of an input image before using them for object detection.
We describe the relevant details in Section~\ref{sec:trans}. Before that, let us discuss details on how to
retrieve the scores from the most similar scene layout cluster given a query image.

\begin{figure}[t]
	\begin{center}
		\includegraphics[width=0.49\textwidth]{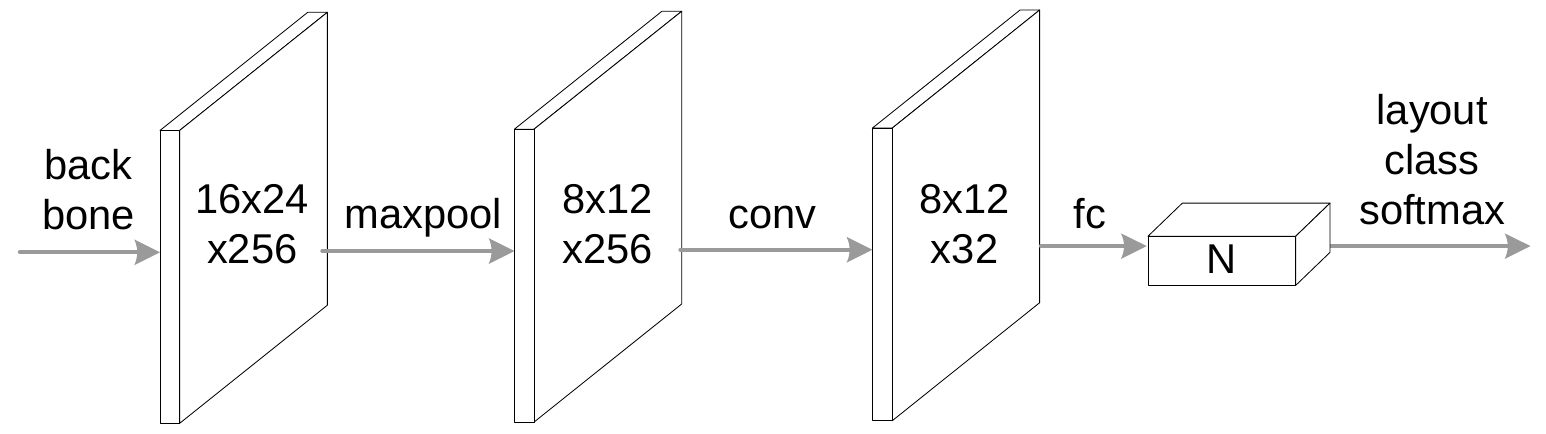}
	\end{center}
	\vspace{-2mm}
	\caption{The scene layout classification head. We use feature map dimensions on the MIO-TCD dataset as
	an example. A max pooling layer and a convolution layer are first applied
	to the feature backbone. Afterwards, a fully connected layer with $N$-way outputs predicts the scene layout cluster
	probabilities. $N$ is the number of clusters in the scene layout codebook.}
	\label{fig:class-head}
	\vspace{-2mm}
\end{figure}

\subsection{Scene layout classification}
\label{sec:cls}

Now we move on to describe the first essential step towards scene layout estimation: scene layout classification.
After building the scene layout codebook with $N$ clusters, we are able to use the cluster membership as a class label for a training image.
Mathematically, let $\mathcal{I}$ be the input image space and $\mathcal{Z} = \{1, \cdots, N\}$ be the label space.
Let $z_m$ be the cluster membership (also, the scene layout class label) for the $m$-th image in the training set.
Therefore, we are given a classification dataset $\{ (I_m, z_m) \}_{m=1}^{M}$ where each $(I_m, z_m) \in (\mathcal{I} \times \mathcal{Z})$.
Our scene layout classifier maps the input image space to the label space $f_{\text{cls}}: \mathcal{I} \rightarrow \mathbb{R}^N$.
We do so by reusing the backbone network of Faster RCNN, and adding an additional sub-network with a softmax output layer.
The sharing of feature computation allows us to avoid unnecessary computational overheads.
Importantly, we note that the scene layout classification allows us to retrieve the most relevant scene layouts in our
codebook, thus making the downstream scene layout transformation and refinement easier to learn.

In terms of the network architecture, we attach a scene layout classification head to the backbone network,
as shown in Figure~\ref{fig:class-head}.
For simplicity in illustrations, we assume that the input image dimension is $512 \times 768$, as we
experimented with the MIO-TCD~\cite{luo2018mio} dataset.
Denote the output of the last residual blocks of $\mathit{conv2}$, $\mathit{conv3}$, $\mathit{conv4}$ and $\mathit{conv5}$
as $\{C_2,C_3,C_4,C_5\}$. Note that, due to the fully convolutional nature of the backbone, these feature maps have
strides of $\{4, 8, 16, 32\}$ pixels with respect to the input image. For instance, the spatial dimensions of $C_5$
would be $16 \times 24\times 256$ for a $512\times 768$ input image, where $256$ is the number of channels.
We use $C_5$ as the input to our scene layout classification head.
In order to obtain a more concise representation, our classification head starts with a $2 \times 2$ max pooling layer and a
$3 \times 3$ stride $1$ convolution layer, followed by a fully connected layer that integrates spatial information for classification.

\begin{figure}[tp!]
	\begin{center}
		\includegraphics[width=0.24\textwidth]{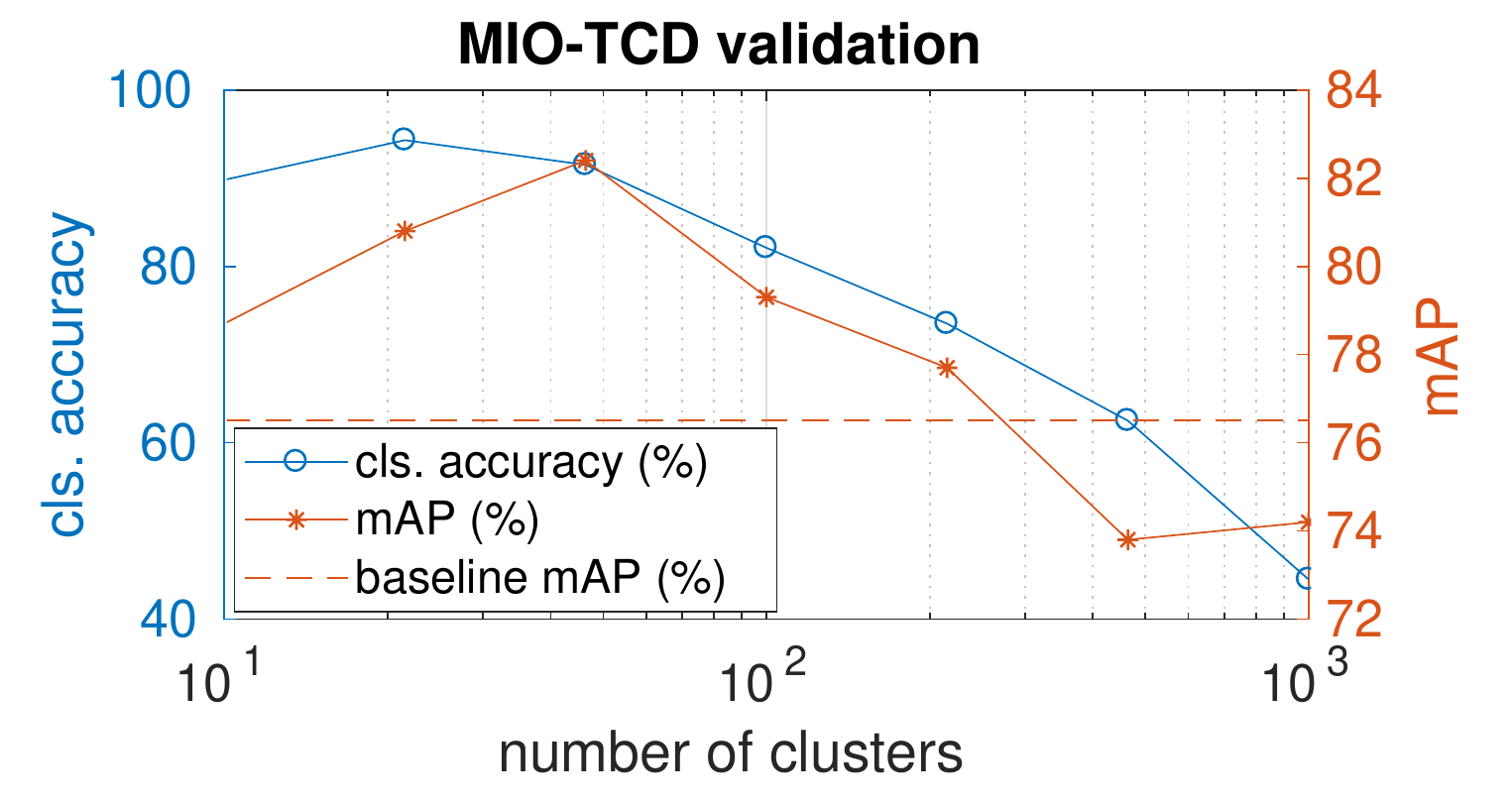}
		\includegraphics[width=0.24\textwidth]{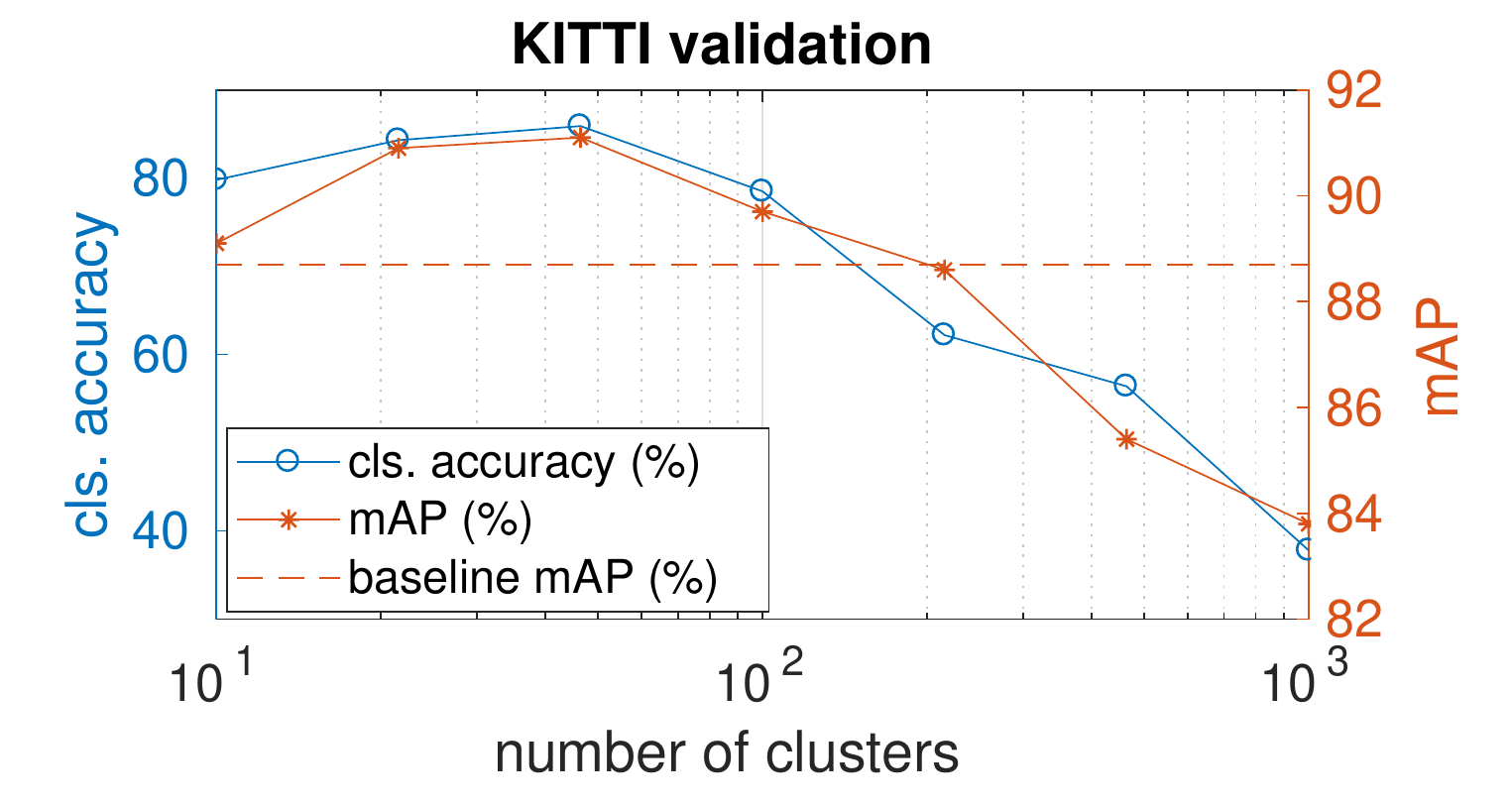}
	\end{center}
	\caption{The impact of number of clusters $N$ on the scene layout classification accuracies and the mean APs.
	The left panel presents results on the MIO-TCD validation set, and the right panel presents results on the
	KITTI validation set.}
	\label{fig:num-clusters}
\end{figure}

We empirically found that setting the number of clusters $N$ properly is important for obtaining the desired scene layout classification results,
as well as object detection performance improvements. Figure~\ref{fig:num-clusters} presents its impact on scene layout
classification accuracies and mean APs for object detection on MIO-TCD and KITTI~\cite{geiger2013vision} validation sets, respectively.
As we can see, if $N$ is too small, the mean AP improvements
may not be maximized and the classification is difficult due to the fact that very different scene layouts may be clustered
into a same scene layout class. On the other hand, setting $N$ to a large value can negatively impact both the classification accuracy
and the mean AP.
In general, we found that $N$ is closely related to the number of typical scene layouts in a dataset.
We set $N=50$ for our experiments in this paper.

\subsection{Scene layout transformation}
\label{sec:trans}

\begin{figure*}[ht!]
	\centering
	\begin{overpic}[scale=.53]%
		{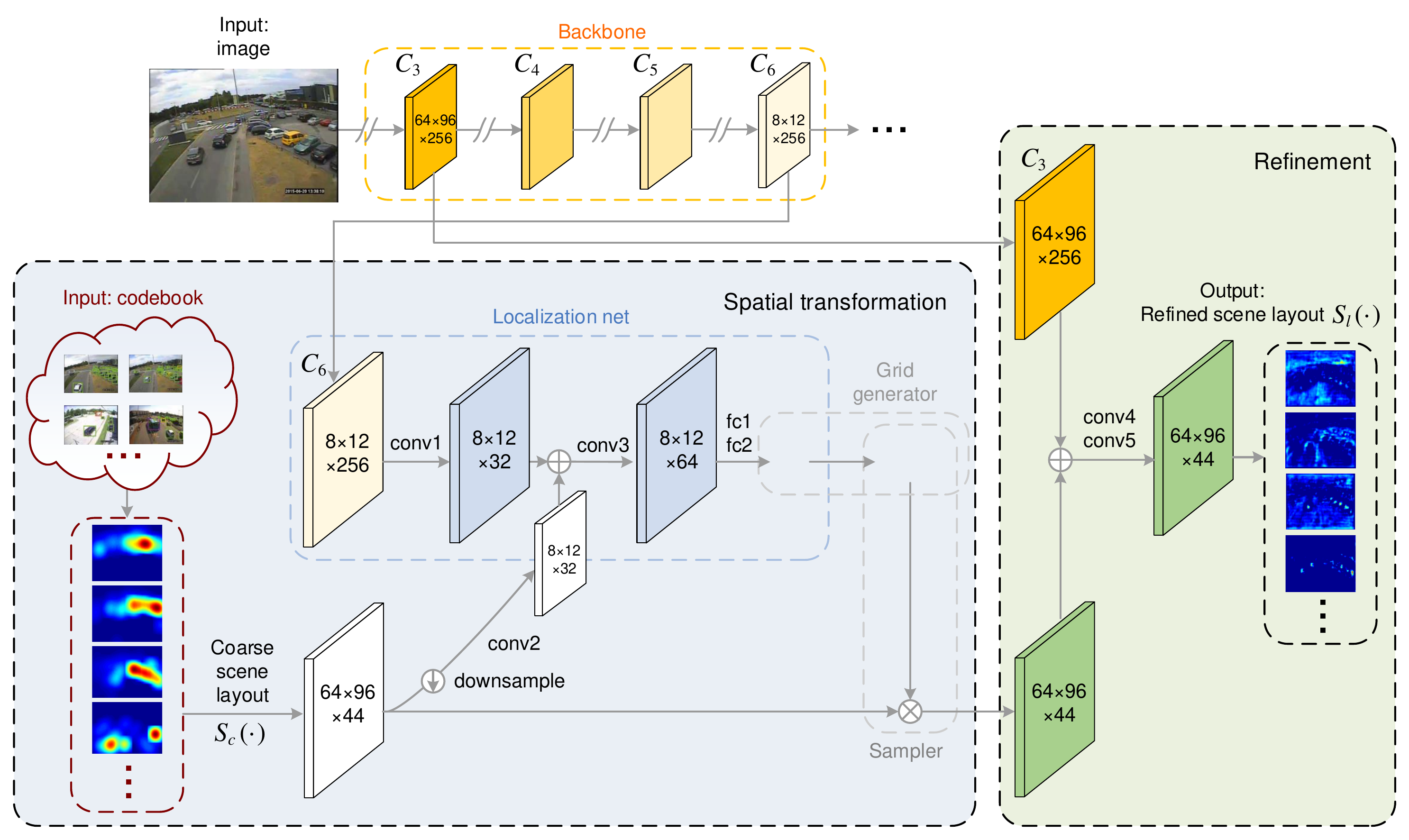}
		\put(55.5,26.2){${\bm \theta}$}
		\put(62.0,26.2){$\mathcal{T}_{\bm \theta}(G)$}
	\end{overpic}
	\caption{The scene layout transformation sub-network. We use the feature map dimensions on the MIO-TCD dataset as an example.
	The sub-network consists of two modules: spatial transformation (the blue panel) and refinement (the green panel).
	Layer parameters are summarized in Table~\ref{tab:stn-params}.
	}
	\label{fig:trans-head}
\end{figure*}

As discussed in the previous section, we are able to retrieve a scene layout template in terms of object
location heatmaps by matching an input image
against the scene layout codebook.
See Figure~\ref{fig:codebook} for a few examples.
Due to large intra-cluster scene layout variations, however, the retrieved templates are
usually too coarse if directly used as a feature map for object detection. See Section~\ref{sec:exp-miotcd} for a detailed
quantitative analysis.
In order to handle the large layout variations, we propose a transformation and a refinement sub-network to adapt the scene layout template to the appearance of a specific input image.
In particular, the spatial transformer network (STN)~\cite{jaderberg2015spatial} proposes a generic CNN module that
allows us to learn any feature transformation in a parameterized form, provided that it is differentiable with respect to
the parameters.

Specifically, for an input image $I$ suppose that the $i$-th cluster is the scene layout classification result
(i.e., the $i$-th element in the output of the
scene layout classification head has the largest logit), so that we retrieve the corresponding scene layout score $S_c^i(\bx|\mathcal{C})$.
We propose to use a variant of the STN to apply a transformation $\mathcal{T}$ to $S_c(\cdot)$ to
obtain the final scene layout score $S_l(\cdot)$,
i.e., $S_l(\bx|\mathcal{N}_I) = \mathcal{T}\big(S^i_c(\bx|\mathcal{C})\big)$.
More specifically, $\mathcal{T}$ consists of two consecutive CNN modules (namely, \textit{spatial transformation} and \textit{refinement},
as shown in Figure~\ref{fig:trans-head}):

\subsubsection{Spatial transformation} Given the coarse scene layout, we first apply a 2D affine transformation
to allow for rotation, translation, scale changes, etc.
Because we use a mixture of discretized groups of scales and aspect ratios in this work, we can reduce the pose of an
object hypothesis $\bx$ to its object center
coordinates $\bx_c=(a_x, a_y)$. Let $(a_x, a_y)$ and $(a'_x, a'_y)$ be the object center coordinates
before and after the spatial transformation, respectively.
In this case, the pointwise transformation in the homogeneous coordinates can be written as:
\begin{align}
	\begin{pmatrix}
	  a_x \\
	  a_y \\
	  1
	\end{pmatrix}
	=
	\begin{bmatrix}
	  \theta_{11} & \theta_{12} & \theta_{13} \\
	  \theta_{21} & \theta_{22} & \theta_{23} \\
	  0           & 0           & 1
	\end{bmatrix}
	\begin{pmatrix}
	  a'_x \\
	  a'_y \\
	  1
	\end{pmatrix}
	\label{eqn:affine}
\end{align}

As we can see from Equation~\ref{eqn:affine}, the 2D affine transformation $\mathcal{T}_{\bm \theta}$ has $6$ parameters.
The STN estimates these parameters with a localization
network $f_{\text{loc}}: (\mathcal{I} \times \mathcal{C}) \rightarrow \mathbb{R}^6$.
In addition, we note that the transformed coordinates $(a'_x, a'_y)$ are defined on a regular grid
$G=\{(a'_x, a'_y)\}$ on an output feature map. Therefore, we can use a sampler to obtain their
corresponding coordinates $\mathcal{T}_{\bm \theta}(G)$
in $(a_x, a_y)$ by applying Equation~\ref{eqn:affine}.

Figure~\ref{fig:trans-head} presents the detailed network architecture of our spatial transformation module.
For simplicity, the feature dimensions on the MIO-TCD dataset are shown as an example.
The dataset has $11$ object categories, and in this paper we use $4$ scale groups and a single aspect ratio group for each object category,
so the number of channels for $S_c(\cdot)$ and $S_l(\cdot)$ is $11 \times 4 \times 1=44$.
At the top of Figure~\ref{fig:trans-head} is the feature backbone, which is shared with downstream object detection modules.
As before, we denote the output of the last residual blocks of $\mathit{conv2}$, $\mathit{conv3}$, $\mathit{conv4}$ and $\mathit{conv5}$
as $\{C_2,C_3,C_4,C_5\}$. Additionally, denote $C_6$ as the feature map that is downsampled by half from $C_5$ with a $2 \times 2$ max pooling layer.
There are two important backbone feature maps we are using in our scene layout transformation sub-network.
The first one is $C_6$, whose spatial dimension is $8 \times 12 \times 256$ and has a stride of $64$ pixels with respect to the input image.
This layer can be used as a feature that is indicative of the overall appearance of images, and with a relatively small number of parameters.
The other one is $C_3$, whose spatial dimension is $64 \times 96 \times 256$ and has a stride of $8$ pixels with respect to the input image.
We will use it as a lower level (i.e., finer) appearance feature that helps our model to attend to specific object locations in the refinement
module.
The spatial resolution of $C_3$ is identical to those object location heatmaps $S_c(\cdot)$ we store with the
codebook $\mathcal{C}$ in our implementation.
We note that, however, this is only a particular implementation choice; there may be other potentially appropriate choices.

The localization network $f_\text{loc}$ is shown in the blue dashed box in Figure~\ref{fig:trans-head}.
In order to regress the localization parameters ${\bm \theta}$, we use both the coarse scene layout $S_c(\cdot)$ and the backbone features $C_6$.
The idea is that the transformation parameters will be learned from both the retrieved (coarse) scene
layout, and the appearance of the input image.
$S_c(\cdot)$ is first downsampled from $64\times 96\times 44$ to $8\times 12\times 44$. The resultant feature maps and $C_6$
are then bottlenecked to $32$ channels (with $\mathit{conv1}$ and $\mathit{conv2}$, respectively) and then concatenated
before going through a convolutional layer ($\mathit{conv3}$) and two fully connected layers ($\mathit{fc1}$ and $\mathit{fc2}$)
in order to regress ${\bm \theta}$.
${\bm \theta}$ is then supplied to the grid generator and the sampler,
which are standard components in an STN~\cite{jaderberg2015spatial}.

\subsubsection{Refinement} \label{sec:refine}
In addition to the affine transformation, we would like the refined scene layout $S_l(\cdot)$ to be able to
have an attention mechanism that focuses on specific regions within a particular input image that are likely to contain objects
(see Figure~\ref{fig:motivation}).
As this would require lower level image features for more accurate localization, we make use of $C_3$,
and concatenate it with the affine-transformed scene layout before going through
two additional convolutional layers ($\mathit{conv4}$ and $\mathit{conv5}$), to obtain the final transformed scene layout $S_l(\cdot)$.
We note that $\mathit{conv4}$ and $\mathit{conv5}$ are different to feature maps $C_4$ and $C_5$ in Figure~\ref{fig:trans-head}
as they learn dedicated feature transformations in order to obtain the refined scene layout $S_l(\cdot)$. On the contrary,
$C_4$ and $C_5$ are the shared backbone features.
The module is shown in the green dashed box in Figure~\ref{fig:trans-head}.
The specific layer parameters in our scene layout transformation sub-network are summarized in Table~\ref{tab:stn-params}.

\begin{table}[t]
	\centering
    \footnotesize
			\begin{tabular}{c c c}
             & layer parameters & output size \\
\shline
             \multicolumn{3}{l}{} \\[-0.9em] % Spacer
             \multicolumn{3}{l}{Spatial transformation} \\[.1em]
\hline
             \multicolumn{3}{l}{} \\[-0.9em] % Spacer
			 $\mathit{conv1}$ & $3 \times 3 \times 32$, stride $1$  &  $8 \times 12 \times 32$ \\
			 $\mathit{conv2}$ & $3 \times 3 \times 32$, stride $1$  &  $8 \times 12 \times 32$ \\
			 $\mathit{conv3}$ & $3 \times 3 \times 64$, stride $1$  &  $8 \times 12 \times 64$ \\	
			 $\mathit{fc1}$ & $6144\rightarrow512$  &  $512$ \\
			 $\mathit{fc2}$ & $512\rightarrow6$  &  $6$ \\
\hline
             \multicolumn{3}{l}{} \\[-0.9em] % Spacer
             \multicolumn{3}{l}{Refinement} \\[.1em]
\hline
             \multicolumn{3}{l}{} \\[-0.9em] % Spacer
			 $\mathit{conv4}$ & $3 \times 3 \times 256$, stride $1$  &  $64 \times 96 \times 256$ \\
			 $\mathit{conv5}$ & $3 \times 3 \times 44$, stride $1$  &  $64 \times 96 \times 44$ \\
         \end{tabular}
         \vspace{2mm}
         \caption
         {
            Layer parameters of the scene layout transformation sub-network used on the MIO-TCD dataset.
            See Figure~\ref{fig:trans-head} for the network architecture.
         }
\label{tab:stn-params}
\vspace{-5mm}
\end{table}

Table~\ref{tab:ltn-ablations} presents some performance comparisons among using different backbone features
for our scene layout transformation sub-network. For the spatial transformation module, we experimented with
$C_6$, $C_4$ and $C_6+C_4$, respectively. In order to use $C_4$ for our localization network, we downsample the
feature map so that it has the same spatial resolution to $C_6$, before going through the downstream processings
as shown in Figure~\ref{fig:trans-head}. We left out $C_5$ because $C_6$ is directly obtained from $C_5$ through max pooling.
The use of $C_4$ allows us to assess the efficacy of lower level features when it comes to estimating the
spatial transformation parameters. It is clear that $C_6$ performs better than the alternatives.
For the refinement module, we tested with $C_3$, $C_3+C_4$ and $C_3+C_4+C_5$, respectively.
$C_4$ and $C_5$ are upsampled where necessary. The idea is to verify
if a cascade of feature maps at multiple feature pyramid levels would additionally benefit the object detection performance.
We empirically found that using $C_3$ alone is sufficient to provide a strong performance.
We note that, however, in this case $C_4$ and $C_5$ are still used for object detection as backbone features.
In general, although the exact connectivity pattern may vary beyond the cases being discussed above, we observe consistent
performance improvements when our layout transformation sub-network is used. See Section~\ref{sec:experiments} for details.

We present some examples from the outputs of our scene layout transformation in Figure~\ref{fig:transformer-output},
which shows the potential locations for the \textit{car} category at $4$ different scales.

\begin{table}[t]
\renewcommand{\arraystretch}{1.3}
	\centering
    \footnotesize
			\begin{tabular}{c | c c c}
             \emph{backbone-features} & AP & AP$_{50}$ & AP$_{75}$\\[.1em]
\shline			
             \multicolumn{3}{l}{Spatial transformation} \\[.1em]
\hline
			 $C_6$ & \bf{57.5} & \bf{80.9} & \bf{64.8} \\
			 $C_4$ & 52.2 & 78.2 & 57.9 \\
			 $C_6+C_4$ & 55.9 & 80.0 & 62.8 \\			 
\hline
             \multicolumn{3}{l}{Refinement} \\[.1em]
\hline
			 $C_3$ & \bf{57.5} & \bf{80.9} & \bf{64.8} \\
			 $C_3+C_4$ & 57.2 & 80.8 & 64.2 \\
			 $C_3+C_4+C_5$ & 56.8 & 80.5 & 63.9 \\			 
         \end{tabular}
         \vspace{2mm}
         \caption
         {
            Average precision values we obtained on the MIO-TCD validation set
            while using various combinations of backbone features in our layout transformation sub-network.
         }
\label{tab:ltn-ablations}
\vspace{-2mm}
\end{table}

\subsection{Piecing things together}
\label{sec:together}

The scene layout predictions, as shown in Figure~\ref{fig:transformer-output}, provide useful context cues for object detection.
In this paper, we propose two strategies to integrate these cues into an object detection framework:

\subsubsection{Late fusion} Following~\cite{wang2017efficient}, we can directly use the scene layout scores
in conjunction with the output of an object detector. The final object detection score is a weighted sum of the two scores:

\begin{align}
	  S(\bx) = S_d(\bx|I)+\alpha S_l(\bx|\mathcal{N}_I)
	  \label{eqn:weighted-sum}
\end{align}

\noindent where $S_d(\bx|I)$ is the scoring function of an object detector, such as Faster RCNN.
$\alpha$ is a hyperparameter for the relative importance between the two terms.

\subsubsection{Early fusion} We use the scene layout scores as an intermediate representation that enhances
our image features. In particular, we experimented with various feature fusion methods, as shown in
Table~\ref{tab:ablation:fusion}. Our best-performing model uses an $1 \times 1$ conv layer,
as illustrated in Figure~\ref{fig:net-struct}.
Specifically, for each level in the feature pyramid~\cite{lin2017feature},
we resize the scene layout scores to the corresponding feature map resolution with bilinear interpolation,
and then concatenate the scores to the feature map.
The $1\x 1$ convolution then maps the features back to its original dimensions before concatenation.
The main advantage of the early fusion strategy is that it results in a model that allows for alternating
optimization of object detection and scene layout estimation.
In Section~\ref{sec:experiments}, we compare early fusion with late fusion on the three datasets
evaluated in this paper.
The details of the training schedule for early fusion are presented
in Section~\ref{sec:impl}.

\subsection{Parameter learning}

In this section, we discuss details pertaining to the learning of the scene layout transfer for object detection.
Particularly, the interpretable nature of our scene layout classification and transformation sub-networks allows
us to inject supervision signals during training. We do so by adding additional terms to the learning
objective of the object detection algorithm. We begin by discussing our overall learning objective.

\subsubsection{Learning objective}

Denote the loss function of our baseline object detection algorithm as $\mathcal{L}_\text{det}$. In the case of
Faster RCNN~\cite{ren2015faster}, this is a multi-task learning objective that involves object classification and bounding box regression.
The overall learning objective of our proposed method can be written as follows:

\begin{align}
	\mathcal{L} = \mathcal{L}_\text{det} + \mathcal{L}_\text{cls} + \mathcal{L}_\text{stn}
\end{align}

\noindent where $\mathcal{L}_\text{cls}$ and $\mathcal{L}_\text{stn}$ are the loss functions for scene layout
classification and transformation, respectively.

\subsubsection{Learning scene layout classification}

For scene layout classification, we use the multi-class cross entropy loss, which is the most widely used loss function
for neural network based classification. Specifically, the scene layout classification loss $\mathcal{L}_\text{cls}$ is defined as:

\begin{align}
      \mathcal{L}_\text{cls} = -\sum_{m=1}^{M}\sum_{i=1}^{N} \delta(z_m = i) \log f_{\text{cls}}^{i}(I_m)
	  \label{eqn:cross-entropy}
\end{align}

\noindent where $\delta(\cdot)$ is the indicator function and $f_{\text{cls}}^{i}(I_m)$ denotes the $i$-th element of $f_{\text{cls}}(I_m)$.

\subsubsection{Learning scene layout transformation}

The output of our scene layout transformation sub-network, $S_l(\bx| \mathcal{N}_I)$, is a set
of heatmaps for object locations of different object categories at various scales and aspect ratios.
We assume that the spatial dimensions of $S_l(\cdot)$ are $W \times H \times K$, where
$W$ and $H$ are the width and height of the heatmaps, and $K=N_{\text{o}} \times N_{\text{s}} \times N_{\text{ar}}$
is the number of components in the mixture model discussed in Section~\ref{sec:codebook},
e.g., $K=11\times 4 \times 1=44$ in the case of the MIO-TCD dataset.
Here $N_{\text{o}}$, $N_{\text{s}}$ and $N_{\text{ar}}$ are the
number of object categories, scale groups, and aspect ratio groups, respectively.
Although it is
possible to train the scene layout transformation sub-network without additional supervision, 
we are able to learn the network parameters with target scene layouts derived from the ground-truth annotations, which we denote as $S^*_l(\cdot)$.
Overall, we write our scene layout transformation loss as follows:

\begin{align}
	\mathcal{L}_\text{stn} = \mathcal{L}_\text{lyt} + \beta \mathcal{L}_\text{reg}
\end{align}

\noindent where the layout loss, $\mathcal{L}_\text{lyt}$, quantifies the mismatch between $S_l(\cdot)$ and $S^*_l(\cdot)$,
and $\mathcal{L}_\text{reg}$ is a regularization term to account for the fact that the affine transformation
shall not drift too far from an identity mapping. Here $\beta$ is a hyperparameter for the tradeoff between
the two terms.

The layout loss $\mathcal{L}_\text{lyt}$ is an element-wise mean squared error (MSE) loss given by:

\begin{align}
	\mathcal{L}_\text{lyt} = \frac{1}{WHK} \sum_{m=1}^{M} \sum_{a_x=1}^{W} \sum_{a_y=1}^{H} \sum_{k=1}^{K} & \big(S_{l,m}^*(\bx | \mathcal{N}_{I_m})_{a_x,a_y,k} \notag \\
	&- S_{l,m}(\bx | \mathcal{N}_{I_m})_{a_x,a_y,k}\big)^2
\end{align}

\noindent Here the target scene layout output of the $m$-th image, $S_{l,m}^*(\bx | \mathcal{N}_{I_m})$, is
obtained by accumulating ground-truth object annotations in that image:

\begin{align}
S_{l,m}^*(\bx | \mathcal{N}_{I_m}) = \frac{1}{Z_l\sigma^2} \sum_{\substack{\bx \simeq \by\\
      \by \in \mathcal{Y}_m}} \exp \Big( - \frac{||\by_c - \bx_c||^2}{2 \sigma^2} \Big)
\end{align}

\noindent where $\by = (\by_c, a_s, a_r, o), \by \in \mathcal{Y}_m$ is a ground-truth object annotation, and $\mathcal{Y}_m$ is the set of all
ground-truth object annotations in the $m$-th image. $\bx \simeq \by$ denotes that $\bx$ and $\by$ are in the same mixture model component.
Here $\bx_c$ and $\by_c$ are
the object center coordinates, and $Z_l$ is a normalizing factor.
Finally, the regularization term $\mathcal{L}_\text{reg}$ is given by:

\begin{align}
	\mathcal{L}_\text{reg} = \sum_{m=1}^{M}\frac{1}{N_{\bm \theta}}||{\bm \theta}^* - {\bm \theta}_m||_2^2
	\label{eqn:layout-reg}
\end{align}

\noindent where ${\bm \theta}_m = [\theta_{11}~\theta_{12}~\theta_{13}~\theta_{21}~\theta_{22}~\theta_{23}]_m^T$ are the parameters
of the affine transformation and
${\bm \theta}^* = [1~0~0~0~1~0]^T$ is the identity transformation. $N_{\bm \theta} = 6$ is the number of elements in ${\bm \theta}$.

\begin{table}[t]
	\centering
    \footnotesize
			\begin{tabular}{c c c c c}
             dataset & input & $C_3$ & $C_6$ & $\mathit{conv3}$ \\
\shline
            \multicolumn{4}{l}{} \\[-0.9em] % Spacer
			MIO-TCD & $512 \times 768$ & $64 \times 96$ & $8 \times 12$ & $3 \times 3 \times 64$ \\
			Traffic lights & $736 \times 1280$ & $92 \times 160$ & $12 \times 20$ & $3 \times 3 \times 32$ \\
			KITTI & $512 \times 1664$ & $64 \times 208$ & $8 \times 26$ & $3 \times 3 \times 32$\\			
         \end{tabular}
         \vspace{2mm}
         \caption
         {
            Feature map resolutions and the kernel dimensions of $\mathit{conv3}$ in the
            transformation sub-network. See Section~\ref{sec:impl} for details.
         }
\label{tab:impl-resolutions}
\vspace{-5mm}
\end{table}

\subsection{Training details}
\label{sec:impl}

In this paper, we adopt a $3$-step training schedule to learn object detection and scene layout
estimation in an alternating fashion.
In the first step, we train the Faster RCNN detector following the standard practice
described in the paper~\cite{ren2015faster}. A scene layout codebook is also learned offline. In the second step, we freeze
the backbone features, and learn the scene layout classification sub-network and the scene layout
transformation sub-network \textit{sequentially}. This order is important because without an accurate
scene layout being retrieved, the training of the transformation sub-network would become unstable.
In this step, we use the SGD solver with initial learning rates set to $0.0025$.
In the third step, we fine-tune the object detector again with the scene layout classification and transformation sub-networks fixed.
The initial learning rate is reduced to $0.001$.
The above alternating training can be run for more iterations, but we observed negligible improvements.

Due to the fully convolutional nature of the backbone network, the dimensions of the feature maps will
depend on the resolution of the input image. The settings being used in this paper are summarized
in Table~\ref{tab:impl-resolutions}. For the Bosch Small Traffic Lights~\cite{behrendt2017deep} and the KITTI datasets, the
dimensions of $C_6$ are higher than that of the MIO-TCD dataset, so we reduce the number of convolutional
kernels by half in $\mathit{conv3}$ (i.e., $64\rightarrow32$), in order to reduce the number of parameters in the $\mathit{fc1}$ layer
of the transformation sub-network.
We note that, however, that our model is somewhat robust to such changes in the network architecture.
It should also be noted that, the higher input image resolution is essential for the Bosch Small Traffic Lights dataset, as most of the
objects are smaller than $32 \times 32$ pixels. In addition, the start size of RPN anchors are also reduced
from $32$ to $8$ to better cope with small objects.

\section{Experimental Results}
\label{sec:experiments}

In this section, we thoroughly compare the proposed method with state-of-the-art object detection algorithms
on public benchmark datasets,
and present ablation studies to verify the efficacy of our model design choices.
We focus on two important sub-areas in vision processing for intelligent transportation systems: traffic surveillance
and autonomous driving.
We evaluate the proposed method on three challenging object detection datasets:
\begin{itemize}
\item \textbf{MIOvision Traffic Camera Dataset (MIO-TCD)}~\cite{luo2018mio}: To the best of our knowledge, MIO-TCD is the largest public benchmark
for object detection in traffic surveillance images, with $110,000$ images
for training and $27,743$ images for testing for its detection task. For this test set, the dataset offers a public object detection
challenge: the TSWC-2017 localization challenge. It allows participating teams to upload their test results to
an evaluation server to obtain their detection performance. Ranking of entries are based on $\text{AP}_{50}$, following
the PASCAL VOC challenge~\cite{pascal-voc-2012}.
\item \textbf{Bosch Small Traffic Lights Dataset}~\cite{behrendt2017deep}: The Bosch Small Traffic Lights dataset presents a unique challenge for detecting small objects with partial occlusions. 
The training and testing tests contain $5,093$ and $8,334$ images, respectively. Context information can be
particularly useful for localizing small objects when the appearance cues are weak from objects themselves.
\item \textbf{KITTI Object Detection}~\cite{geiger2013vision}: The KITTI object detection dataset has $7,481$ training images
and $7,518$ testing images.
As our method crucially relies on the availability of large-scale datasets that cover objects at various locations and scales,
we evaluate the detection performance of the most prevalent \textit{car} category that has $28,742$ training instances.
The KITTI dataset also offers a public online benchmark based on $\text{AP}_{70}$.
\end{itemize}

We refer our method to Layout Transfer Network (LTN) in the following sections.

\subsection{Results on the MIO-TCD dataset}
\label{sec:exp-miotcd}

\begin{figure*}[ht!]
	  \centering
	\includegraphics[width=0.9\textwidth]{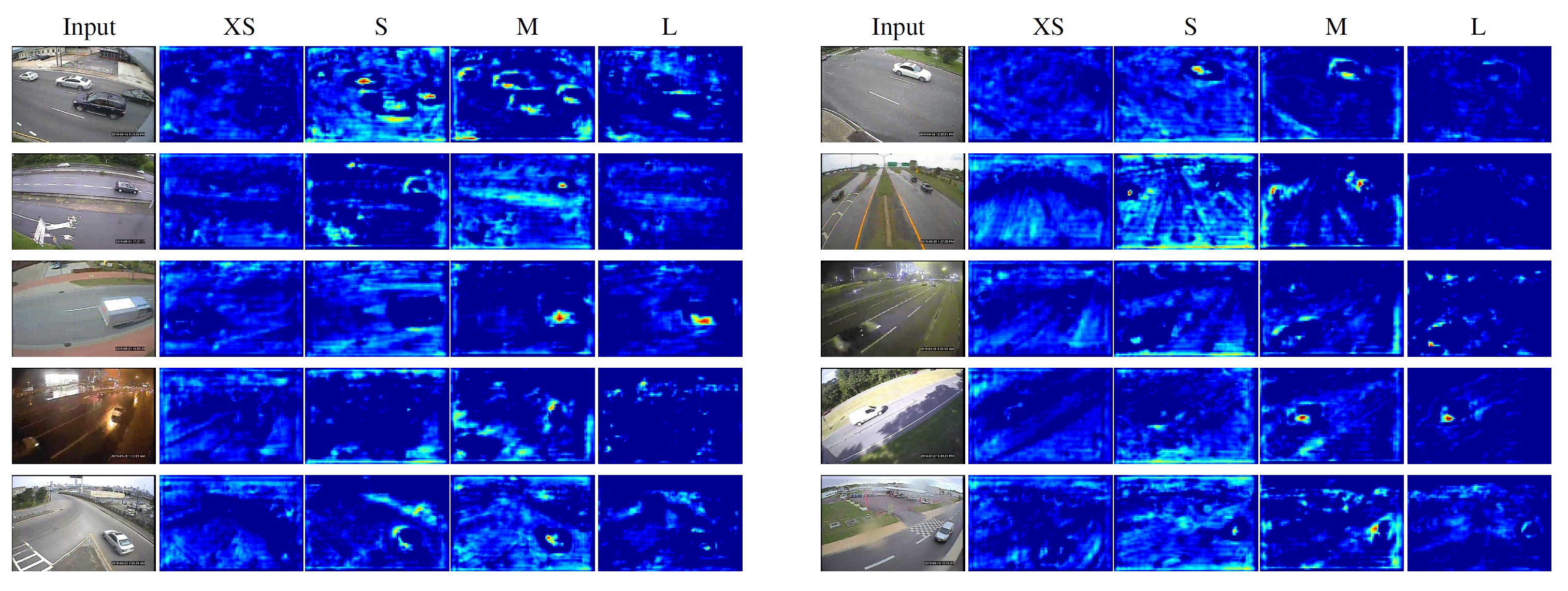}
	  
\vspace{1mm}
		\caption
         {
         	Examples of the refined scene layout scores $S_l(\cdot)$ for the \textit{car} category obtained with the MIO-TCD dataset.
         	Input images are shown in the
         	leftmost column, with locations for extra small (XS, farthest), small (S, far), medium (M, close)
         	and large (L, closest) objects in the four columns to the right.
         }
	\label{fig:transformer-output}
\end{figure*}

\begin{table*}[t]
	\renewcommand{\arraystretch}{1.3}
	\centering
    \footnotesize
		\begin{tabular}{l| l| l l l l l l l l l l l | l }

             \emph{MIO-TCD-test} & \scriptsize backbone & \rotatebox{90}{a.truck} & \rotatebox{90}{bicyle} & \rotatebox{90}{bus} &
             \rotatebox{90}{car} & \rotatebox{90}{motorcycle} & \rotatebox{90}{m.vehicle} &
             \rotatebox{90}{n.m.vehicle} & \rotatebox{90}{pedestrian} & \rotatebox{90}{p.truck} &
             \rotatebox{90}{s.u.truck} & \rotatebox{90}{workvan} & \rotatebox{90}{mean $\text{AP}_{50}$} \\[.2em]
             \shline
%             \multicolumn{10}{l}{} \\[-0.9em] % Spacer
             \multicolumn{10}{l}{Challenge submissions} \\[.1em]
             \hline
%             \multicolumn{10}{l}{} \\[-0.9em] % Spacer
             \textbf{LTN~(Ours)} & \scriptsize R101-FPN & 92.38 & \bf{88.42} & \bf{97.64} & \bf{95.14} & \bf{92.65} & \bf{70.65} & 58.41 & \bf{67.31} &
             92.74 & \bf{74.97} & \bf{79.93} & \bf{82.75} \\                          
             Jung et al.~\cite{jung2017resnet} & \scriptsize R101-ensemble & \bf{92.48} & 87.34 & 97.46 & 89.70 & 88.21 & 62.32 & \bf{59.09} & 48.57 &
             92.25 & 74.42 & 79.86 & 79.24 \\
             Wang et al.~\cite{wang2017efficient} & \scriptsize VGG-16+context & 91.62 & 79.90 & 96.77 & 93.80 & 83.63 & 56.40 & 58.23 & 42.61 &
             \bf{92.75} & 73.80 & 79.56 & 77.19 \\
%             \multicolumn{10}{l}{} \\[-0.9em] % Spacer             
             \hline
 %            \multicolumn{10}{l}{} \\[-0.9em] % Spacer             
             \multicolumn{10}{l}{Baseline approaches} \\[.1em]
             \hline
%             \multicolumn{10}{l}{} \\[-0.9em] % Spacer             
             SSD-512~\cite{liu2016ssd} & \scriptsize VGG-16 & 91.28 & 77.36 & 96.56 & 93.59 & 79.53 & 55.39 & 56.60 & 41.58 &
             92.66 & 72.74 & 79.40 & 76.06 \\
             YOLO v2~\cite{redmon2017yolo9000} & \scriptsize DarkNet-19 & 88.31 & 78.64 & 95.13 & 81.36 & 81.36 & 51.70 & 56.57 & 24.96 &
             86.48 & 69.23 & 76.43 & 71.83 \\             
             Faster RCNN~\cite{ren2015faster} & \scriptsize VGG-16 & 80.70 & 70.63 & 93.45 & 79.85 & 74.58 & 46.48 & 21.22 & 19.49 &
             86.71 & 53.29 & 67.40 & 63.07 \\
         \end{tabular}
         \vspace{2mm}
         \caption
         {
            Per-class and mean average precision values~($\text{AP}_{50}$, in $\%$) we obtained in the TSWC-2017 localization
            challenge (MIO-TCD test set).
            Our entry obtained a mean $\text{AP}_{50}$ of $82.75$, which is the state-of-the-art. We have the highest $\text{AP}_{50}$ in $8$ out
            of $11$ object categories among all entries.
         }
\label{tab:mio-ap-test}

\end{table*}

%##################################################################################################
\begin{table*}[t]\vspace{-3mm}
% subfloat a - early late fusion
\subfloat[\textbf{Early \vs late fusion}: We compare two fusion methods for the learned scene layout scores.
Early fusion allows for joint training for object detection and scene layout estimation, and compares favorably to late fusion.\label{tab:ablation:earlylate}]{
\tablestyle{4pt}{1.3}\begin{tabular}{c|x{22}x{22}x{22}}
 \scriptsize{\emph{early-vs-late-fusion}} & AP & AP$_{50}$ & AP$_{75}$\\[.1em]
\shline
 \scriptsize LTN~--~late fusion & 55.2 & 79.5 & 61.9\\
 \scriptsize LTN~--~early fusion & \bd{57.5} & \bd{80.9} & \bd{64.8}\\
\hline
 \scriptsize & \dt{+2.3} & \dt{+1.4} & \dt{+2.9}\\
% \multicolumn{4}{c}{~}\\
\end{tabular}}\hspace{10mm}
% subfloat b - fusion method
\subfloat[\textbf{Fusion method}: For early feature-level fusion, we experimented with different network
structures. $1 \times 1$ conv outperforms other alternatives.\label{tab:ablation:fusion}]{
\tablestyle{2.5pt}{1.3}\begin{tabular}{c|x{22}x{22}x{22}}
 \scriptsize \emph{fusion-method} & AP & AP$_{50}$ & AP$_{75}$\\[.1em]
\shline
 \scriptsize eltwise-mul & 56.2 & 80.0 & 63.3\\
 \scriptsize eltwise-sum & 54.7 & 79.5 & 61.2\\
% \scriptsize scaled dot-product & 55.6 & 79.7 & 62.5\\
 \scriptsize $1\x1$ conv & \bd{57.5} & \bd{80.9} & \bd{64.8}\\
\end{tabular}}
\hspace{10mm}
% subfloat c - layout resolution
\subfloat[\textbf{Layout resolution:} AP results with various layout resolutions. Resolutions above $64\x96$
(i.e., the resolution of the $C_3$ feature map) give diminishing returns.
%We use $64\x96$ (or the resolution of the $C_3$ feature map) unless otherwise specified.
\label{tab:ablation:layout-res}]{
\tablestyle{4pt}{1.3}\begin{tabular}{c|x{22}x{22}x{22}}
 \scriptsize{\emph{layout-resolution}} & AP & AP$_{50}$ & AP$_{75}$ \\[.1em]
\shline
 \scriptsize $32\x48$ & 55.4 & 79.8 & 61.9 \\
 \scriptsize $64\x96$ & 57.5 & 80.9 & 64.8 \\
 \scriptsize $128\x192$ & 57.7 & 80.9 & 65.1\\
% \multicolumn{4}{c}{~}\\
\end{tabular}}
\\
% subfloat d - component
\subfloat[\textbf{Component test}: To isolate
the performance gains, we switch off the following three components from our full model:
(1) scene layout classification, (2) spatial transformation, and (3) refinement.
Each component provides its own AP improvements, while the full model performs the best.
\label{tab:ablation:component}]{
\tablestyle{2.2pt}{1.3}\begin{tabular}{c|c|c|c|x{22}x{22}x{22}}
 & \scriptsize classify? & \scriptsize transform? & \scriptsize refine?
 & AP & AP$_{50}$ & AP$_{75}$\\
\shline
 \scriptsize Baseline
  & \xmark & \xmark & \xmark & 48.2 & 76.1 & 52.7\\
\hline
 \multirow{3}{*}{\scriptsize +~components}
  & \cmark & \xmark & \xmark & 52.0 & 77.6 & 58.0\\
  & \cmark & \cmark & \xmark & 53.3 & 78.8 & 59.6\\
  & \cmark & \xmark & \cmark & 55.5 & 79.8 & 62.2\\
\hline
 \scriptsize Full model & \cmark & \cmark & \cmark & \bd{57.5} & \bd{80.9} & \bd{64.8}
\end{tabular}}
\hspace{5mm}
% subfloat e - FCN vs LTN
\subfloat[\textbf{Fully convolutional network (FCN) \vs Layout transfer network (LTN):} Directly predicting scene layouts with an FCN does not produce superior results.\label{tab:ablation:fcnltn}]{
\tablestyle{4.8pt}{1.3}\begin{tabular}{c|x{22}x{22}x{22}}
 & AP & AP$_{50}$ & AP$_{75}$\\[.1em]
\shline
 FCN & 51.1 & 76.2 & 58.1\\
 LTN & \bd{57.5} & \bd{80.9} & \bd{64.8}\\
\hline
 & \dt{+6.4} & \dt{+4.7} & \dt{+6.7}\\
 \multicolumn{4}{c}{~}\\
 \multicolumn{4}{c}{~}\\
\end{tabular}}
\hspace{5mm}
% subfloat f - layout regularization
\subfloat[\textbf{Layout regularization}: The layout regularization term is essential for learning the spatial
transformation. We observe large performance gaps without the regularizer being used.\label{tab:ablation:layout-reg}]{
\tablestyle{4pt}{1.3}\begin{tabular}{c|x{22}x{22}x{22}}
 & AP & AP$_{50}$ & AP$_{75}$ \\[.1em]
\shline
 \scriptsize w/o $\mathcal{L}_{\text{reg}}$ & 53.0 & 78.6 & 58.6 \\
 \scriptsize w/ $\mathcal{L}_{\text{reg}}$ & \bd{57.5} & \bd{80.9} & \bd{64.8} \\
\hline
 & \dt{+4.5} & \dt{+2.3} & \dt{+6.2}\\
 \multicolumn{4}{c}{~}\\
 \multicolumn{4}{c}{~}\\
\end{tabular}}\vspace{2mm}
% main caption
\caption{\textbf{Ablations}. We report average precision values from our ablation studies on the MIO-TCD validation set.
The backbone network is a ResNet-101-FPN~\cite{lin2017feature}. See Section~\ref{sec:exp-miotcd} for discussions.}
\label{tab:ablations}\vspace{-3mm}
\end{table*}
%##################################################################################################

In order to perform detailed ablation studies, we randomly split the original training set of the MIO-TCD dataset into
a smaller training set with $99,000$ images and a held-out validation set with $11,000$ images. All results reported in
this section are obtained by training models on the smaller training set, with ablation studies being carried out on the
held-out validation set.

The quantitative results we obtained in the TSWC-2017 localization challenge\footnote{http://podoce.dinf.usherbrooke.ca/results/localization}
(i.e., the MIO-TCD test set) are reported in Table~\ref{tab:mio-ap-test}.
Comparing to other entries, our method shows a clear advantage. In particular, our results are $3.51$
points better in terms of $\text{AP}_{50}$ than the current winning entry from Jung et al.~\cite{jung2017resnet}. It should be noted, however,
that their method uses a 4-models ensemble (two ResNet-50s and two ResNet-101s) based on R-FCN~\cite{dai2016r} and ours use only a single model. In addition, our
results are $5.56$ $\text{AP}_{50}$ points better than our previous work~\cite{wang2017efficient} that uses a nonparametric label transfer
method to predict the scene layout for object detection. Overall, we obtain the highest average precisions on $8$ out of $11$ object categories.
To the best of our knowledge, these results are the
state-of-the-art in the MIO-TCD localization challenge.

In addition to the results on the test set, we also report some ablations we obtained on the
held-out validation set, which are presented in Table~\ref{tab:ablations}. Specifically, we perform the following
ablation studies:

\begin{itemize}
\item \textbf{Early \vs late fusion} (Table~\ref{tab:ablation:earlylate}): We compare the performance of early and late
fusion as discussed in Section~\ref{sec:together}. As expected, the early fusion strategy allows us to jointly optimize
for object detection and scene layout estimation, resulting in better detection performance.

\item \textbf{Fusion method} (Table~\ref{tab:ablation:fusion}): In the case of early fusion, there are various ways to
integrate the inferred scene layout features $S_l(\cdot)$ with backbone image features. In this set of experiments, we
explore different ways to feature fusion, including elementwise multiplication (eltwise-mul), elementwise summation (eltwise-sum),
and $1 \x 1$ convolution. The $1 \x 1$ convolution variant performs best and is used in all other experiments by default.
We note that, for the two elementwise fusion methods, we need to apply an $1 \x 1$ convolution layer before
the elementwise operations to make sure that the feature dimensions of the backbone and the scene layout are compatible.

\item \textbf{Layout resolution} (Table~\ref{tab:ablation:layout-res}): Because we use a pyramid of features in the FPN
backbone, a higher scene layout resolution could be helpful for detecting and localizing smaller objects.
In general, a layout
resolution at $64 \x 96$ for the MIO-TCD dataset (i.e., the resolution of $C_3$) is sufficient and higher resolutions
give marginal performance gains.
In this work, we stick to the resolution of $C_3$ as the scene layout resolution on all three datasets, as it provides
a good tradeoff between performance and model complexity (i.e., the number of parameters).

\item \textbf{Component test} (Table~\ref{tab:ablation:component}): The three main modules in the scene layout estimation
are: (1) scene layout classification, (2) spatial transformation, and (3) refinement. Compared to a baseline Faster RCNN
that uses none of these components, adding each module or any combination of them contribute to the AP performance.

\item \textbf{FCN \vs LTN} (Table~\ref{tab:ablation:fcnltn}): One of the main advantages of our proposed method is that
we are able to retrieve a rough estimate of the scene layout in the first step, making the subsequent transformation and refinement
easier to learn. Here we also report results obtained by learning a fully convolutional network that directly predicts the
scene layout features $S_l(\cdot)$ from the backbone features $C_3$ and $C_6$.
As we can see, our retrieve-and-transform strategy boosts the
detection performance by a reasonable margin.

\item \textbf{Layout regularization} (Table~\ref{tab:ablation:layout-reg}): Finally, we verify the effectiveness of our
layout regularization term in Equation~\ref{eqn:layout-reg}. We observe a large performance improvement with the regularizer
being used. In addition, we found in our experiments that the regularization term helps stabilize training and avoid over-fitting.
\end{itemize}

\begin{table}[t]
\renewcommand{\arraystretch}{1.3}
	\centering
    \footnotesize
			\begin{tabular}{c | c c c}
             \emph{Traffic-lights-test} & AP & AP$_{50}$ & AP$_{75}$\\[.1em]
\shline
			 Faster RCNN & 30.25 & 73.44 & 15.91 \\
			 Faster RCNN~+~LTN (late fusion) & 31.20 & 75.83 & 18.52 \\			 
			 Faster RCNN~+~LTN (early fusion) & \bf{31.64} & \bf{75.97} & \bf{19.06} \\
\hline
			 ~+~LTN (late fusion) & \dt{+1.0} & \dt{+2.4} & \dt{+2.6}\\
			 ~+~LTN (early fusion) & \dt{+1.4} & \dt{+2.5} & \dt{+3.2}\\
         \end{tabular}
         \vspace{2mm}
         \caption
         {
            Average precision values we obtained on the Bosch Small Traffic Lights test set.
         }
\label{tab:tl-ap-test}
\vspace{-2mm}
\end{table}

\begin{figure*}[t]
	  \centering
      \includegraphics[width=0.7\textwidth]{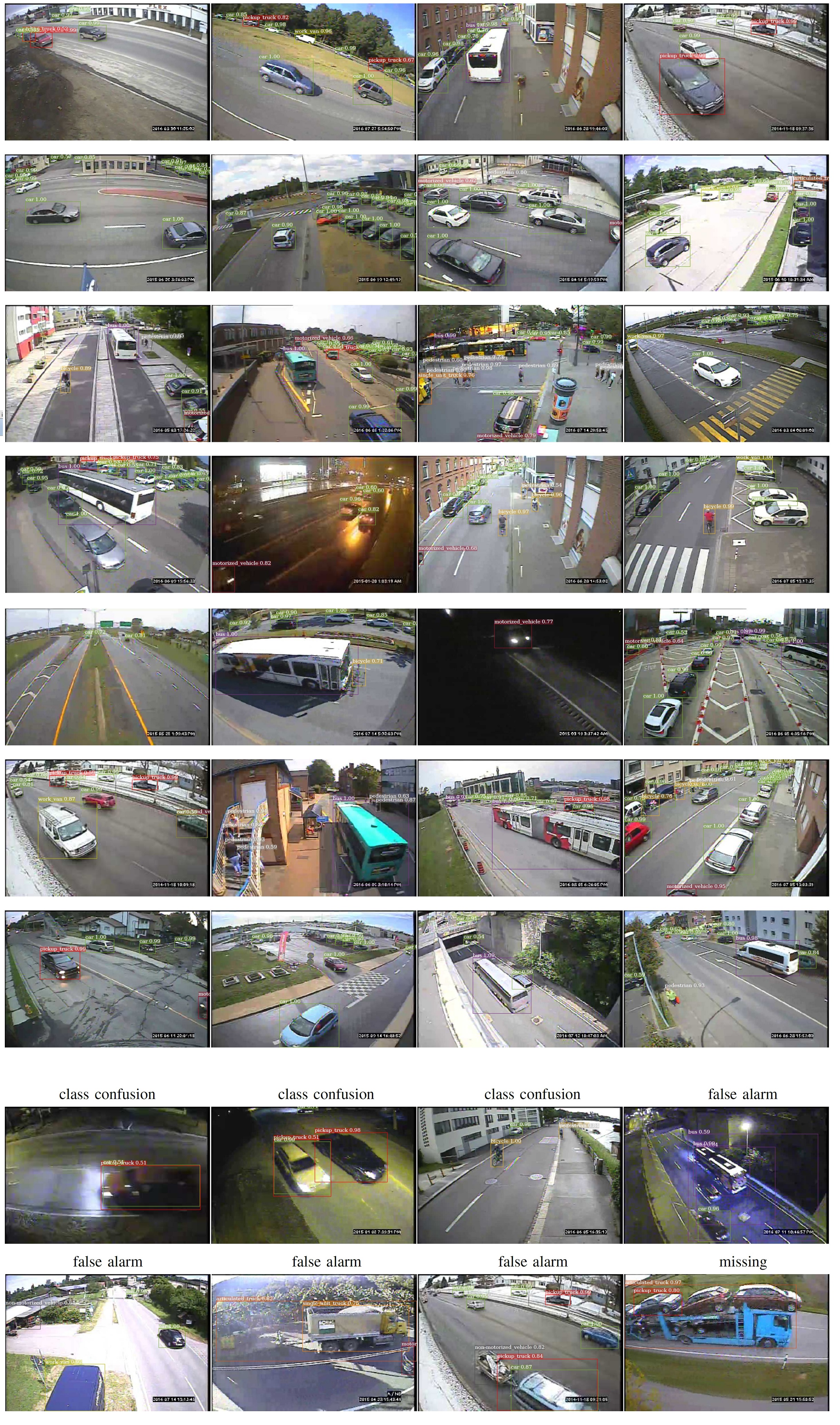}
		\caption
         {
			Example detection results on the MIO-TCD test set. The last two rows present some typical failure
			modes, namely class confusion, false alarm, and missing out-of-context objects.
			See Section~\ref{sec:exp-miotcd} for details. Best viewed electronically, zoomed in.
         }
      \label{fig:detections-miotcd}
\end{figure*}

It is aforementioned that our method provides an interpretable representation of the scene layouts.
In Figure~\ref{fig:transformer-output}, we present some examples of the scene layouts for the \textit{car} category
at $4$ different scales as predicted by our method.
It is clear from these examples that our scene layout scores can reliably predict object locations and scales through
the object location heatmaps.
In addition, we present some qualitative detection results
in Figure~\ref{fig:detections-miotcd}. As we can see, our method is able to reliably detect overlapping and distant
objects by incorporating the context cues. Typical failure modes are also presented in the last two rows,
among the most common are class confusions, false alarms, and missing out-of-context objects.

\begin{figure*}[t]
	  \centering
      \includegraphics[width=0.7\textwidth]{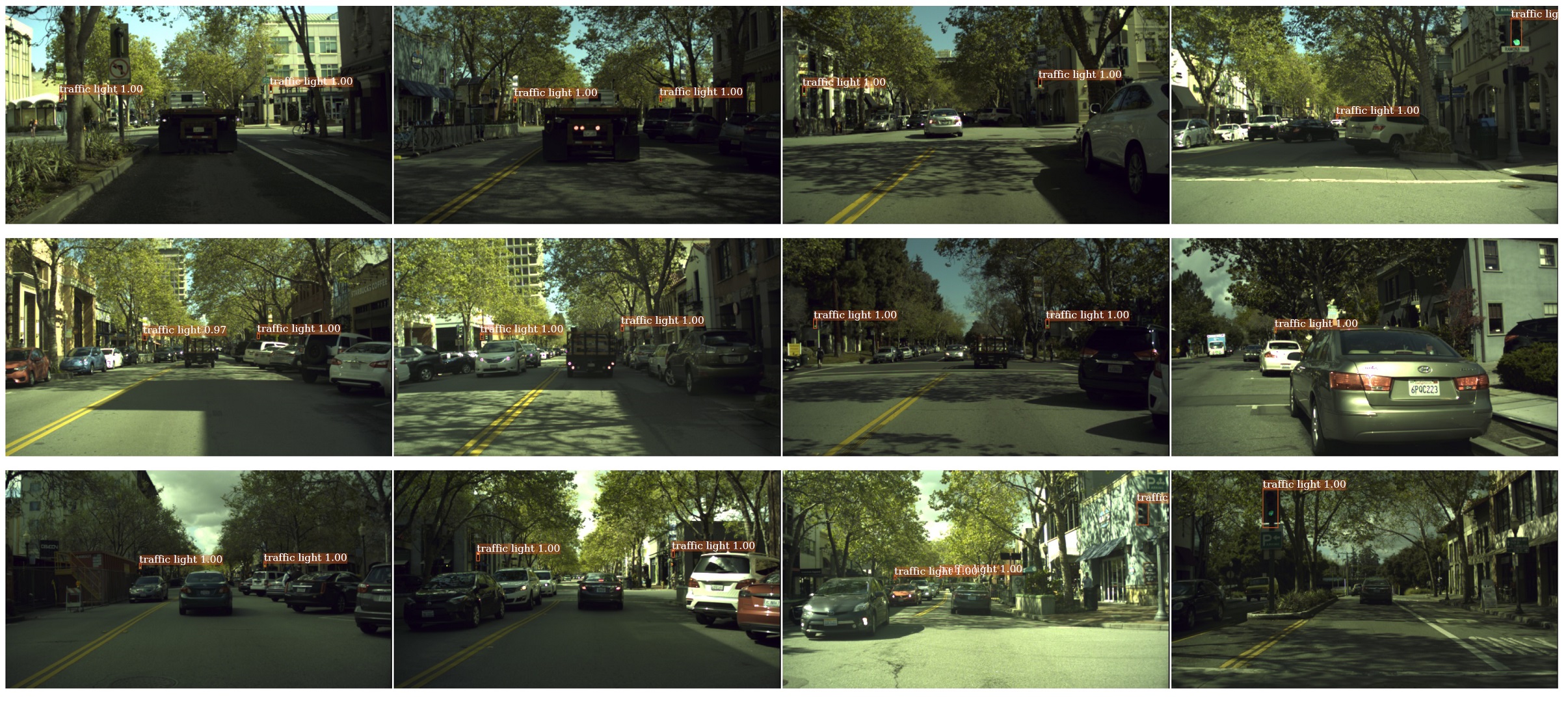}
		\caption
         {
			Example detection results on the Bosch Small Traffic Lights test set. See Section~\ref{sec:exp-tl} for details.
			Best viewed electronically, zoomed in.	
         }
      \label{fig:detections-tl}
\end{figure*}

\begin{figure*}[t]
	  \centering
      \includegraphics[width=0.7\textwidth]{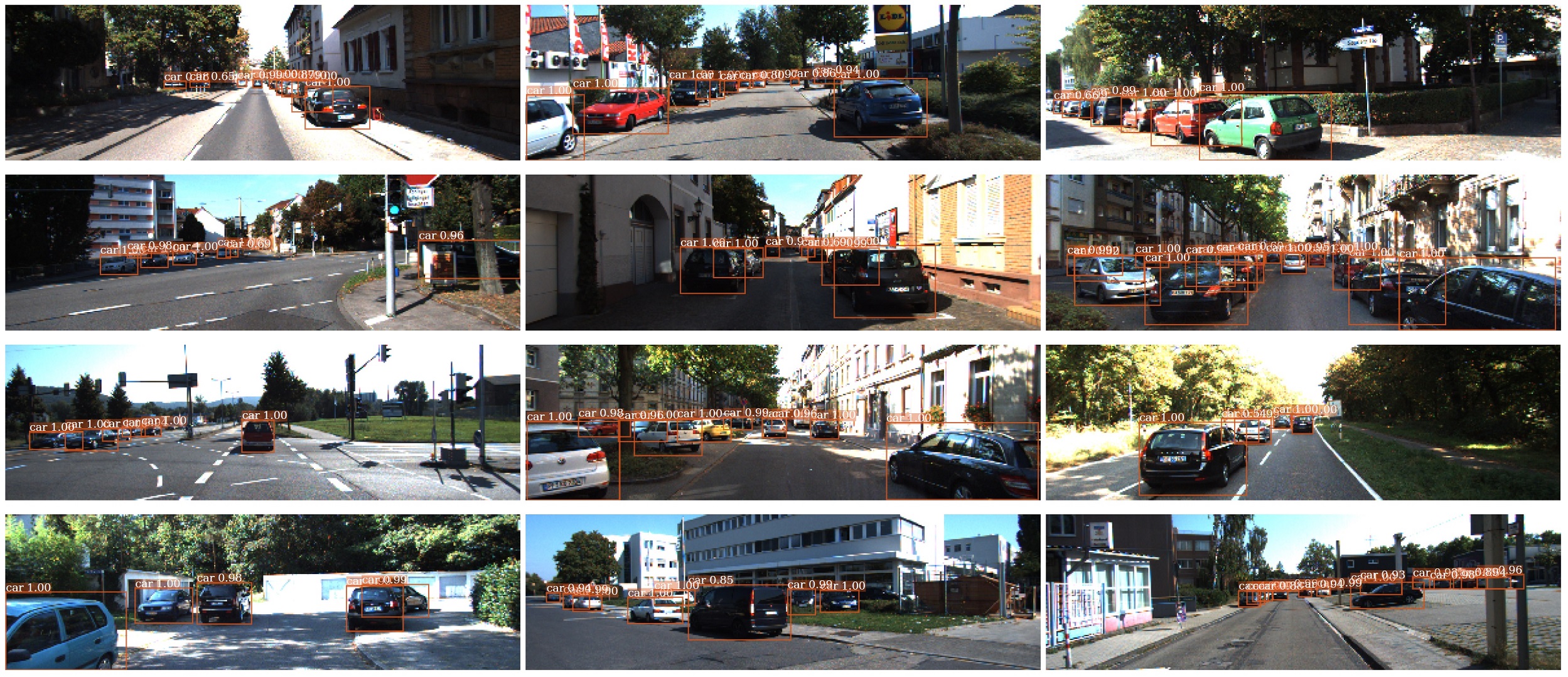}
		\caption
         {
			Example detection results on the KITTI test set. See Section~\ref{sec:exp-kitti} for details.
			Best viewed electronically, zoomed in.
         }
      \label{fig:detections-kitti}
\end{figure*}

\subsection{Results on the Bosch Small Traffic Lights dataset}
\label{sec:exp-tl}

In addition, we report the results we obtained on the Bosch Small Traffic Lights dataset. This dataset presents a unique
challenge to detecting small objects with partial occlusions in autonomous driving.
The detection results we obtained on the test set are reported in Table~\ref{tab:tl-ap-test}.
In particular, accurate localization
for objects as small as a few pixels wide (the medium width is about $8.6$ pixels) is very challenging for state-of-the-art object detectors, as reflected by the low $\text{AP}_{75}$ values in a sharp contrast to the $\text{AP}_{50}$ values.
Although we found that temporal information may be used to improve accuracy on this dataset (as some traffic lights
may be missed while being detected in preceding or succeeding frames), it is not the focus of this paper and is therefore not used. 
Once again, our layout transfer network is able to improve detection results in these very challenging scenarios.
Also, we emphasize again that using high resolution input images,
as specified in Table~\ref{tab:impl-resolutions}, and smaller RPN anchors are vital to obtain satisfactory results on this particular dataset.
We present some example detection results in Figure~\ref{fig:detections-tl}.

\subsection{Results on the KITTI dataset}
\label{sec:exp-kitti}

The KITTI dataset provides a comprehensive set of real-world and challenging computer vision tasks including stereo, optical flow, visual odometry,
object detection and tracking for scene understanding in autonomous driving.
A 2D object detection dataset with $7,481$ annotated training images is provided. As it is vital for our method to
transfer scene layouts from a large dataset with objects at various locations and scales, we only evaluate the detection
performance of \textit{cars}, which is the largest object category.
Other object categories (i.e., \textit{pedestrians} and \textit{cyclists}) are much smaller in comparison,
so we do not include them in our evaluation.
In addition, we split the original training set into a separate training set
and a validation set with $5,985$ and $1,496$ images, respectively. This allows us to directly demonstrate the
effectiveness of LTN, as compared to a baseline with identical settings otherwise.

\begin{table}[t]
\renewcommand{\arraystretch}{1.3}
	\centering
    \footnotesize
			\begin{tabular}{c | c c c}
             \emph{KITTI-val} & Moderate & Easy & Hard \\[.1em]
\shline
			 Faster RCNN & 88.72 & 90.76 & 80.31 \\
 			 Faster RCNN~+~LTN (late fusion) & \bf{93.12} & 94.04 & 86.25 \\
			 Faster RCNN~+~LTN (early fusion) & 93.09 & \bf{94.42} & \bf{86.65} \\
\hline
			 ~+~LTN (late fusion) & \dt{+4.4} & \dt{+3.3} & \dt{+5.9}\\
			 ~+~LTN (early fusion) & \dt{+4.4} & \dt{+3.7} & \dt{+6.3}\\
         \end{tabular}
         \vspace{2mm}
         \caption
         {
            Average precision values~($\text{AP}_{70}$, in $\%$) we obtained on the KITTI validation set.
         }
\label{tab:kitti-ap-val}
\vspace{-5mm}
\end{table}

Following the evaluation protocol of the dataset, the $\text{AP}_{70}$ values are reported on \textit{Moderate}, \textit{Easy}, and
\textit{Hard} objects, respectively.
Table~\ref{tab:kitti-ap-val} summarizes the results on the KITTI validation set.
Compared to the baseline, our LTN (with early fusion) improves the average precision by $4.4$, $3.7$ and $6.3$ points
on the three difficulty levels, respectively.
In particular, the performance on the \textit{Hard} difficulty level is boosted by a large margin, as contextual cues
are important for detecting objects that are heavily occluded or truncated by image boundaries.
In addition, Table~\ref{tab:kitti-ap-test} reports the results from the KITTI benchmark (i.e., test set).
In general, our method achieves a good tradeoff between accuracy and speed, especially at the \textit{Hard} difficulty level.
In particular, the top-performing method
RRC~\cite{ren2017accurate} is around $10$ times slower than our method due to the recurrent
nature of its architecture.
Another state-of-the-art method, SINet~\cite{hu2018sinet}, is faster than our approach but we perform better at the \textit{Hard}
difficulty level.
In addition, we note that the contributions of others may be
orthogonal to ours.
Furthermore, some state-of-the-art methods on the online
benchmark~\footnote{http://www.cvlibs.net/datasets/kitti} are not listed here because they either require additional input
modalities (e.g., PC-CNN-V2~\cite{du2018general} and F-PointNet~\cite{qi2018frustum})
or CAD models during training (e.g., Deep MANTA~\cite{chabot2017deep}).

We present some qualitative results in Figure~\ref{fig:detections-kitti}.
As we can see, scene layouts can be well indicative of potential vehicle locations and scales, and that our method is able to detect
overlapping and distant objects with high accuracy.

\begin{table} [ht!]
	\renewcommand{\arraystretch}{1.3}

	\centering
	\begin{tabular}{c|c|c|c|c}
		\multirow{2}{*}{\emph{KITTI-test}} &
		\multirow{2}{*}{time/image}  &
		\multicolumn{3}{c}{AP$_{70}$}   \\
		\cline{3-5}  & & Moderate & Easy & Hard   \\
		\shline
		RRC~\cite{ren2017accurate} &	3.6s&	\textbf{90.23} &	90.61&	\textbf{87.44} \\
		SJTU-HW~\cite{zhang2018led,fang2018small} & 0.85s & 90.08 & 90.81 & 79.98 \\
		SINet~\cite{hu2018sinet} &	0.2s&	89.60&	90.60&	77.75\\
		Deep3DBox~\cite{mousavian20173d}&	1.5s&	89.04&	\textbf{92.98}&	77.17 \\
		
		MS-CNN~\cite{cai2016unified}&	0.4s&	89.02&	90.03&	76.11 \\
		Mono3D~\cite{chen2016monocular}&	4.2s&	88.66&	92.33&	78.96 \\
		SDP+CRC (ft)~\cite{yang2016exploit}&	0.6s&	83.53&	90.33&	71.13 \\
		spLBP~\cite{hu2016fast}&	1.5s&	77.40&	87.19&	60.60 \\
		Reinspect~\cite{stewart2016end}&	2s&	76.65&	88.13&	66.23 \\
		Regionlets~\cite{long2014accurate,wang2015regionlets,zou2014generic}&	1s&	76.45&	84.75&	59.70 \\
		SubCat~\cite{ohn2015learning}&	0.7s&	75.46&	84.14&	59.71 \\
\hline
		\textbf{LTN~(Ours)} & 0.4s & 88.85 & 90.12 & 79.62 \\		
	\end{tabular}
	\vspace{2mm}
	\caption{Average precision values~($\text{AP}_{70}$, in $\%$) on the KITTI benchmark.
	Methods are ranked based on their performance at the Moderate
	difficulty level.
	In general, the LTN achieves a good tradeoff between accuracy and speed, especially at
	the Hard difficulty level. See text for details.}
	\label{tab:kitti-ap-test}	
	\vspace{-5mm}
\end{table}

\section{Conclusion}
\label{sec:conclusion}
In this paper, we proposed a layout transfer network for context aware object detection.
An important aspect of our method is that we are able to obtain an interpretable scene layout representation
which can be directly used to improve object detection performance.
The scene layout transfer in our method provides a general approach to context modeling for object detection
that can be used in conjunction with many other detection algorithms not mentioned in this paper.
In the future, we wish to achieve scene layout classification with an integrated
deep learning approach.
In particular, this may allow for integrated training of all sub-systems and parameters, which
may provide better overall performance.
For example, we can use backpropagation to fine-tune the scene layout codebook after the initialization by clustering.
We hope that our work would serve as a modest spur to induce further exploration into simple and robust scene layout
representations that may be useful for a wider variety of scene understanding problems.

% or


%\appendices
%% you can choose not to have a title for an appendix
%% if you want by leaving the argument blank
%\section{}

\section*{Acknowledgment}
\small{
We thank the anonymous reviewers for their insightful comments.
We also thank Zhiming Luo and Pierre-Marc Jodoin for their help in our participation in
the TSWC-2017 challenge. %The GPUs used in this research were donated by NVIDIA Corporation.
Project sponsored by NSFC (61703195, 61702431), Fujian NSF (2019J01756), Shanghai NSF (18ZR1425100),
The Education Department of Fujian Province (JAT170459, JK2017039, and the Distinguished Young Scholars Program),
Fuzhou Technology Planning Program (2018-G-96, 2018-G-98) and Minjiang University (MJUKF201716, MJY19021, MJY19022).
}

\ifCLASSOPTIONcaptionsoff
  \newpage
\fi



%\bibliographystyle{IEEEtran}

\bibliographystyle{IEEEtran}
\bibliography{IEEEabrv,wangbib}


\begin{IEEEbiography}[{\includegraphics[width=1in,height=1.25in,clip,keepaspectratio]{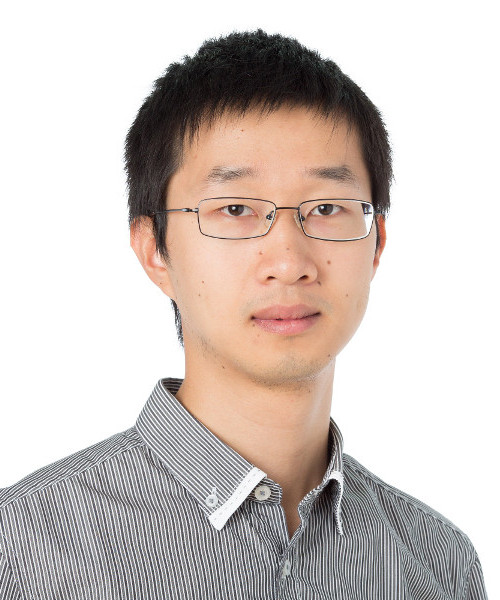}}]{Tao Wang}
received the B.E. degree in information engineering from South China University of Technology, Guangzhou, China, in 2009,
and the Ph.D. degree in computer science from The Australian National University, Canberra, ACT, Australia in 2016.
He was also a member of the computer vision research group at National ICT Australia, Canberra.
He is currently a lecturer with the College of Computer and Control Engineering, Minjiang University, Fuzhou, China.
He is also with
the College of Mathematics and Computer Science,
Fuzhou University, Fuzhou, and NetDragon Inc., Fuzhou.
His research interests include scene understanding, object detection, and semantic instance segmentation.
\end{IEEEbiography}

\begin{IEEEbiography}[{\includegraphics[width=1in,height=1.25in,clip,keepaspectratio]{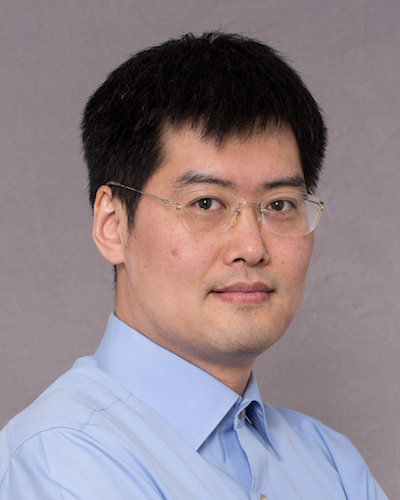}}]{Xuming He}
received the Ph.D. degree in computer science from the University of Toronto, Toronto,
ON, Canada, in 2008.
He held a post-doctoral position with the University
of California at Los Angeles, Los Angeles,
CA, USA, from 2008 to 2010. He was an Adjunct
Research Fellow with The Australian National University,
Canberra, ACT, Australia, from 2010 to
2016. He joined National ICT Australia, Canberra,
where he was a Senior Researcher from 2013 to
2016. He is currently an Associate Professor with
the School of Information Science and Technology, ShanghaiTech University,
Shanghai, China. His research interests include semantic image and video
segmentation, 3-D scene understanding, visual motion analysis, and efficient
inference and learning in structured models.
\end{IEEEbiography}

\begin{IEEEbiography}[{\includegraphics[width=1in,height=1.25in,clip,keepaspectratio]{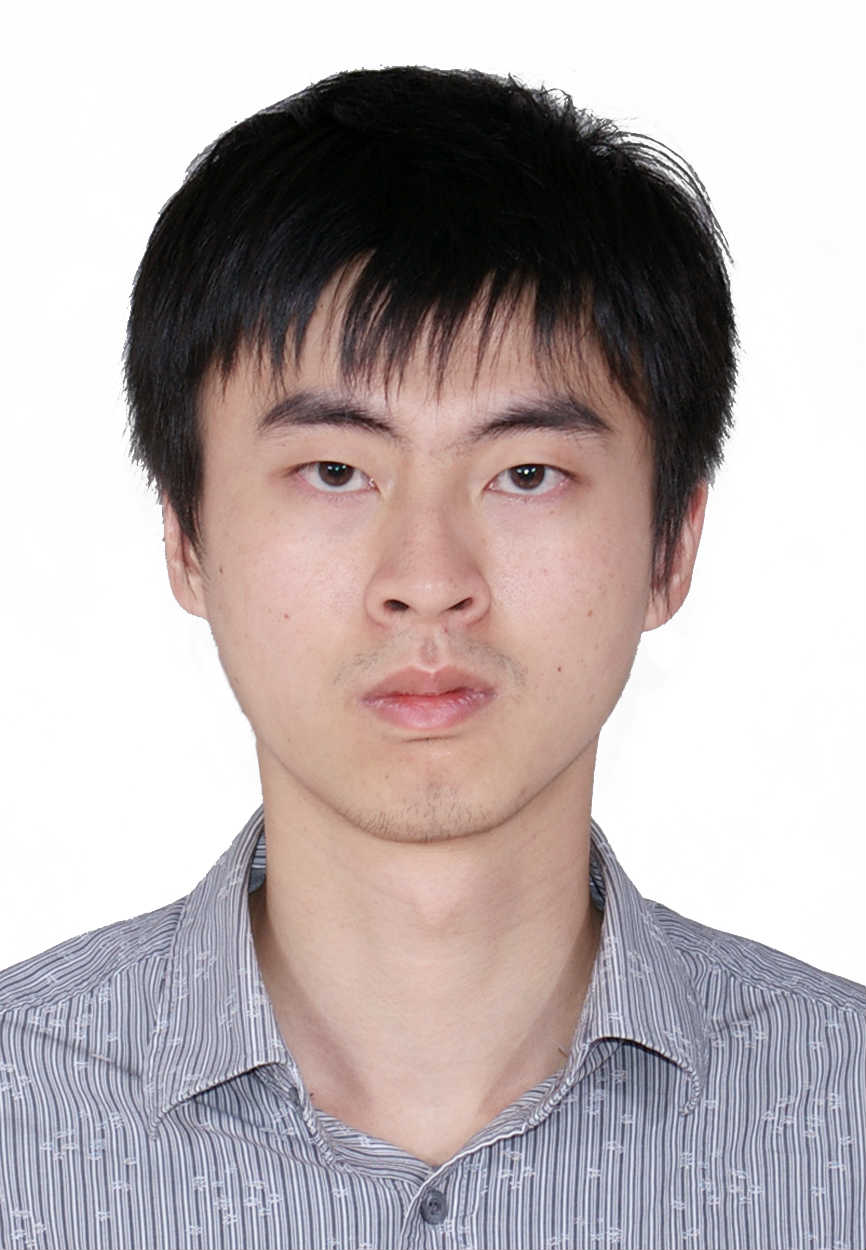}}]{Yuanzheng Cai}
received the B.S. degree in software engineering from Fujian Normal University, Fuzhou, China, in 2010,
the M.S. degree in computer science from Yunnan University, Kunming, China, in 2012, and the Ph.D.
degree in artificial intelligence from Xiamen University, Xiamen, China, in 2016. He is currently a 
lecturer with the College of Computer and Control Engineering, Minjiang University, Fuzhou, China.
He is also with
the College of Mathematics and Computer Science,
Fuzhou University, Fuzhou, and NetDragon Inc., Fuzhou.
His research interests include image/video retrieval,
machine learning, and object recognition.
\end{IEEEbiography}

\begin{IEEEbiography}[{\includegraphics[width=1in,height=1.25in,clip,keepaspectratio]{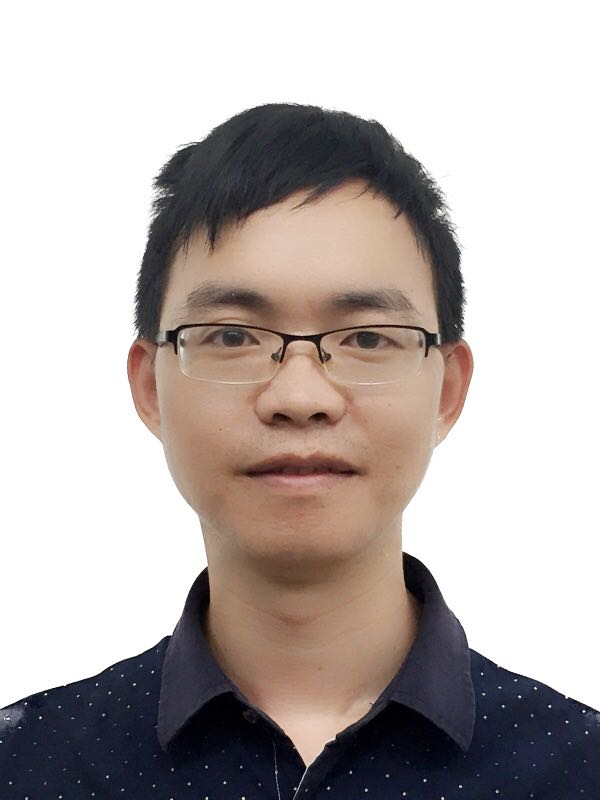}}]{Guobao Xiao}
is currently a Professor at Minjiang University, China. He was a Postdoctoral Fellow (2016-2018) in the School of Aerospace Engineering at Xiamen University, China. He received the Ph.D. degree in Computer Science and Technology from Xiamen University, China, in 2016. He has published over 20 papers in international journals and conferences including IEEE Transactions on Pattern Analysis and Machine Intelligence, International Journal of Computer Vision, Pattern Recognition, Pattern Recognition Letters, Computer Vision and Image Understanding, ICCV, ECCV, ACCV, AAAI, ICIP, ICARCV, etc. His research interests include machine learning, computer vision and pattern recognition.
\end{IEEEbiography}


%\vfill


\end{document}